\documentclass[twocolumn]{svjour3}

\usepackage{multirow}
\usepackage{appendix}
\usepackage[normalem]{ulem}

\usepackage{xcolor}
\usepackage{amsmath}
\DeclareMathOperator*{\argmax}{argmax}
\usepackage{amsfonts}
\usepackage{booktabs}
\usepackage{tabularx,booktabs}

\usepackage{caption}
\usepackage{subcaption}
\usepackage{comment}
\usepackage{array}
\usepackage{graphicx}
\usepackage{nicefrac}
\usepackage{xspace}
\usepackage{relsize}
\usepackage{lmodern}

\usepackage{pgfplots}
\usepackage{pgfplotstable}
\usepackage{pgf,tikz}
\usetikzlibrary{shapes,arrows}
\usetikzlibrary{positioning}
\usetikzlibrary{decorations.pathreplacing}
\usetikzlibrary{decorations.markings}

\pgfplotsset{compat=newest}
\usetikzlibrary{plotmarks}
\usetikzlibrary{patterns}
\usepgfplotslibrary{patchplots}
\pgfplotsset{
	tick label style={font=\scriptsize},
	label style={font=\scriptsize},
	legend style={font=\scriptsize},
	title style={font=\scriptsize}}

\usepackage{scrextend}
\usepackage{hyperref}
\usepackage{bigfoot}
\usepackage{url}

\usepackage{etoolbox,siunitx}
\usepackage{adjustbox}
\usepackage{tabularx}
\robustify\bfseries

\def\addlegendimage{\csname pgfplots@addlegendimage\endcsname}

\newcolumntype{L}{>{$}l<{$}}
\newcolumntype{C}{>{$}c<{$}}
\newcolumntype{R}{>{$}r<{$}}

\renewcommand\footnotemark{}

\makeatletter
\ifcase \@ptsize \relax
  \newcommand{\miniscule}{\@setfontsize\miniscule{4}{5}}
\or
  \newcommand{\miniscule}{\@setfontsize\miniscule{5}{6}}
\or
  \newcommand{\miniscule}{\@setfontsize\miniscule{5}{6}}
\fi
\makeatother

\def\real{\mathbb{R}}

\def\l2{\ensuremath{\ell_2}\xspace}

\renewcommand{\paragraph}[1]{\vspace{.0\baselineskip}\noindent{\bf #1}\xspace}

\def\etal{\emph{et al.}\xspace}
\def\ie{\emph{i.e.}\xspace}
\def\eg{\emph{e.g.}\xspace}

\def\wrt{\emph{w.r.t.}\xspace}

\renewcommand{\paragraph}[1]{{\medskip \noindent \bf #1}}

\usepackage{algorithm}
\usepackage{algpseudocode}

\usepackage{url}            
\usepackage{booktabs}       
\usepackage{cite}
\usepackage{amsmath}
\usepackage{amssymb}
\usepackage{xspace}
\usepackage{nicefrac}
\usepackage{soul}        
\usepackage{lipsum}   
\usepackage{wrapfig}
\usepackage{setspace}
\usepackage{xcolor}
\usepackage{tabularx}
\usepackage{multirow}
\usepackage{makecell}
\usepackage{epsfig}
\usepackage{bm}

\makeatletter
\DeclareRobustCommand\onedot{\futurelet\@let@token\@onedot}
\def\@onedot{\ifx\@let@token.\else.\null\fi\xspace}
\def\eg{\emph{e.g}\onedot} 
\def\ie{\emph{i.e}\onedot} 
\def\cf{\emph{c.f}\onedot} 
 \def\vs{\emph{vs}\onedot} 
\def\wrt{\emph{w.r.t}\onedot} 
\def\etal{\emph{et al}\onedot}
\makeatother

\def\be{\begin{equation}}
\def\ee{\end{equation}}
\def\bea{\begin{eqnarray}}
\def\eea{\end{eqnarray}}

\def\tab#1{Table~\ref{tab:#1}}

\begin{document}

\title{HSCNet++: Hierarchical Scene Coordinate Classification and Regression for Visual Localization with Transformer}

\author{Shuzhe Wang \and Zakaria Laskar \and Iaroslav Melekhov \and Xiaotian Li \and Yi Zhao \and Giorgos Tolias \and Juho Kannala}

\institute{
S. Wang, I. Melekhov, X. Li, Y. Zhao, J. Kannala \at Aalto University in Finland
\\\email{{firstname.lastname@aalto.fi}} \and
Z. Laskar, G. Tolias \at Visual Recognition Group, Faculty of Electrical Engineering, CTU in Prague
\\\email{{laskazak,toliageo@fel.cvut.cz}}
}

\maketitle
\thispagestyle{empty}

\begin{abstract}
Visual localization is critical to many applications in computer vision and robotics. To address single-image RGB localization, state-of-the-art feature-based methods match local descriptors between a query image and a pre-built 3D model. Recently, deep neural networks have been exploited to regress the mapping between raw pixels and 3D coordinates in the scene, and thus the matching is implicitly performed by the forward pass through the network. However, in a large and ambiguous environment, learning such a regression task directly can be difficult for a single network. In this work, we present a new hierarchical scene coordinate network to predict pixel scene coordinates in a coarse-to-fine manner from a single RGB image.
The proposed method, which is an extension of HSCNet,  
allows us to train compact models which scale robustly to large environments. It sets a new state-of-the-art for single-image localization on the 7-Scenes, 12-Scenes, Cambridge Landmarks datasets, and the combined indoor scenes. 
\end{abstract}
\section{Introduction}
\label{sec:intro}
\begin{figure*}[t]
\begin{center}
\includegraphics[width=0.85\textwidth]{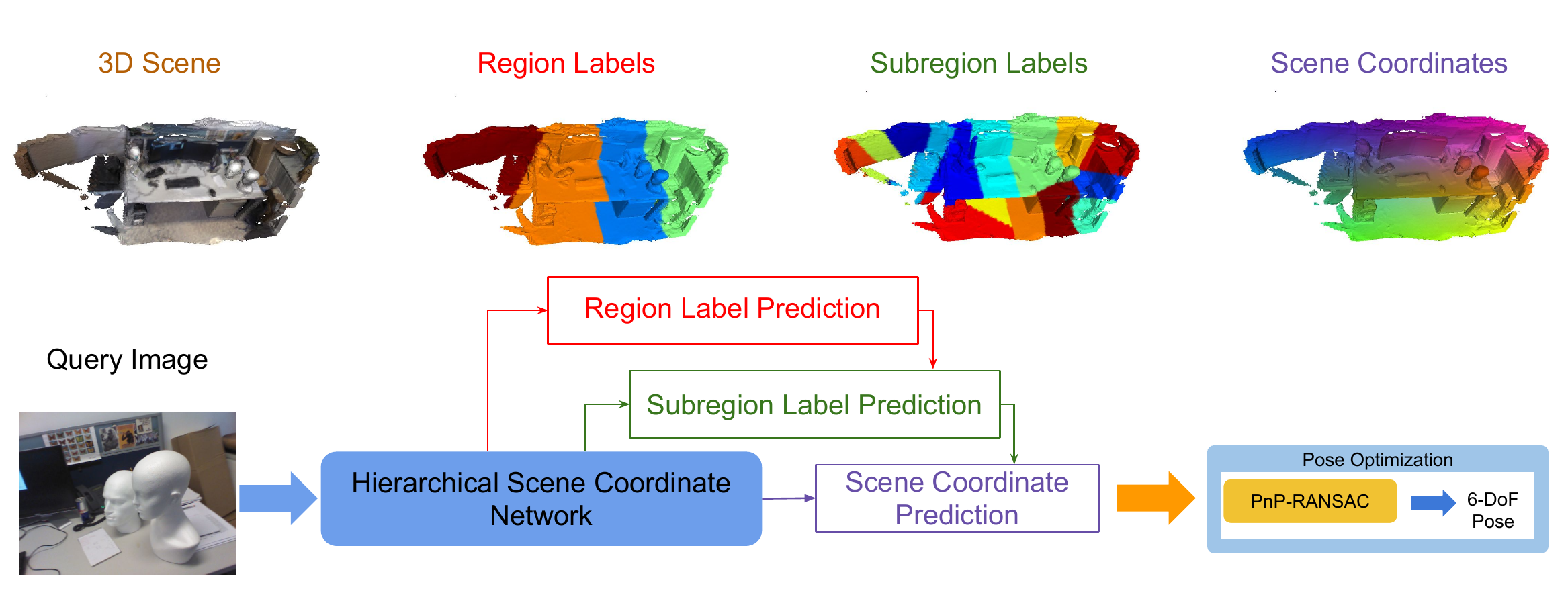}
\end{center}
\caption{
\textbf{HSCNet architecture.} 
The ground-truth scene 3D coordinates are hierarchical quantized into regions and sub-regions. Dirrent branches of the the network sequentially predicts discrete regions and sub-regions, and continous 3D coordinates, with the processing of each branch being conditioned on the result of the previous one.  Given an input image, HSCNet predicts 3D coordinates for 2D image pixels, which then form the input to PnP-RANSAC for 6DoF pose estimation.
}
\label{fig:overview}
\end{figure*}

Estimating the six degrees-of-freedom (6-DoF) camera pose from a given RGB image is a key component in many computer vision systems such as augmented reality, autonomous driving, and robotics. Classical methods~\cite{sattler2011fast, sattler2012improving, sattler2016efficient, taira2018inloc, sarlin2019coarse} establish 2D-2D(-3D) correspondences between query and database local descriptors, followed by PnP-based camera pose estimation. This incurs both storage and computational costs by necessitating storage of millions of database local descriptors and hierarchical descriptor matching in a RANSAC loop.

On the other hand, end-to-end pose regression methods that directly regress the camera pose parameters are much faster and memory efficient~\cite{kendall2015convolutional,Balntas_2018_ECCV, Chen2021DirectPoseNet, shavit2022camera}. However, such methods are significantly less accurate than local descriptor ones. A better trade-off between accuracy and computational efficiency is offered by structured localization approaches~\cite{Brachmann_2017_CVPR, Brachmann_2018_CVPR, brachmann2020visual, SCoRF,li2020hscnet, wang2021continual}. Structured methods are trained to learn an implicit representation of the 3D environment by directly regressing 3D scene coordinates corresponding to a 2D pixel location in a given input image. This directly provides 2D-3D correspondences and avoids storage and explicit matching of database local descriptors with the query. For small-scale scenes, it is shown that scene-coordinate methods~\cite{Brachmann_2018_CVPR,brachmann2020visual} outperform classical local descriptor-based methods, but later~\cite{brachmann2021limits} show that the performance is indeed comparable. Nevertheless, the storage and computational benefits of structured-based methods are superior to classical local descriptor matching methods.

Existing scene-coordinate regression methods~\cite{Brachmann_2017_CVPR,Brachmann_2018_CVPR,brachmann2020visual} are designed to predict scene coordinates from a small local image patch that provides robustness to viewpoint changes. On the other hand, such methods are limited in applicability to larger scenes where ambiguity from visually similar local image patches cannot be resolved with a limited receptive field. Using larger receptive field sizes, up to the full image, to regress the coordinates can mitigate the issues from ambiguities by encoding larger context. 
This, however, is shown to be prone to overfitting the larger input patterns in the case of limited training data, even if data augmentation  alleviates this problem to some extent~\cite{Li2018,brachmann2020visual}.

Increasing context by enlarging the receptive field while maintaining local distinctiveness of descriptors or not overfitting is a challenging problem. We address this using a special network architecture, called HSCNet~\cite{li2020hscnet}, which hierarchically encodes scene context using a series of classification layers before making the final coordinate prediction. The overall pipeline is shown in Fig~\ref{fig:overview}. Particularly, the network predicts scene coordinates progressively in a coarse-to-fine manner, where predictions correspond to a region in the scene at the coarse level and coordinate residuals at the finest level. The predictions at each level are conditioned on both descriptors and predictions from the preceding level which we experimentally show is the key component in large scenes. This conditioning leverages FiLM~\cite{film} layers that allow for a gradual increase in the receptive field. 
Instead of leveraging simple CNNs as in HSCNet to encode the descriptors and predictions, this work extends it to utilize the transformer-based~\cite{vaswani2017attention} conditioning mechanism, named HSCNet++, which is more efficient in capturing global
context into local representations through attention and doesn't require heavy conventional layers to enlarge the receptive field. The architecture manages to improve coordinate prediction at all levels, both coarse and fine. We integrate dynamic position information in the form of predicted coarse positional encoding, without the need to learn or construct explicitly position embeddings and show promising results on several benchmarks.

We further extend HSCNet++ by removing the dependency on dense ground truth scene coordinates. Dense coordinates limit the applicability of HSCNet to outdoor scenes. Similar to~\cite{Brachmann_2018_CVPR}, HSCNet addressed the issue of sparse data on Cambridge dataset~\cite{kendall2015convolutional} by using MVS-based densification~\cite{schoenberger2016mvs}. However, these methods either introduce additional noise and are costly to obtain. Directly training HSCNet with sparse supervision leads to a significant performance drop. In HSCNet++, we propose a simple yet effective pseudo-labelling method, where ground-truth labels at each pixel location are propagated to a fixed spatial neighbourhood. This is based on the assumption that nearby pixels share similar statistics. To provide robustness to pseudo-label noise, symmetric loss functions based on cross-entropy and reprojection loss are proposed. While the symmetric cross-entropy loss provides robustness to the classification layers of HSCNet, the reprojection loss rectifies the noise in pseudo-labelled 3D scene coordinates.

This work is a summary and extension of HSCNet. We validate our approach on three datasets used in previous works: %
7-Scenes~\cite{SCoRF}, 12-Scenes~\cite{valentin2016learning}, and Cambridge Landmarks~\cite{kendall2015convolutional}. Our approach shows consistently better performance and achieves state-of-the-art results for single-image RGB localization. In addition, by compiling the 7-Scenes and 12-Scenes datasets into single large scenes,
we show that our approach scales more robustly to larger environments. 
In summary, our contributions  are as follows:
\begin{enumerate}
    \item Compared to HSCNet, we utilize an improved transformer based conditioning mechanism that efficiently and effectively encodes global spatial information to scene coordinate prediction pipeline, resulting in a significant performance improvement from 84.8\% to 88.7\% on indoor localization while requiring only 57\% of the memory footprint. 
    \item We extend HSCNet to optionally leverage the sparse ground truth only in the training procedure by introducing pseudo ground truth labels and angle-based reprojection errors. When using sparse supervision for training, HSCNet++(S) achieves better accuracy on the Cambridge dataset compared to HSCNet++ trained on MVS-densified data.
    \item We show that the classical pixel-based positional encoding in our conditioning mechanism suffers from significant performance drop, especially in scenes which have massive repetitive patterns. Our spatial positional encoding by the FiLM layer eliminates this problem and achieves SoTA performance on several benchmarks.
\end{enumerate}

\section{Related Work}
\label{sec:related}
Existing methods for visual localization are reviewed depending on the category they belong to.

\paragraph{Classical visual localization} methods assume that a scene is represented by a 3D model, which is a result of processing a set of database images. Each 3D point of the model is associated with one or several database local descriptors. Given a query image, a sparse set of keypoints and their local descriptors are obtained using traditional~\cite{Calonder2019Brief,SIFT,Rublee2011ORB,Bay2006SURF} or learned CNN-based~\cite{DeTone_2018_CVPR_Workshops,Revaud2019R2D2,Dusmanu2019CVPR,Melekhov2021hndesc, Melekhov2020Stylization,luo2019contextdesc,Wang2020CAPS,l2net2017Tian,Balntas2016TFeat,Zagoruyko2015DeepCompare,Han2015MatchNet,Melekhov2017PatchMatch,Simo-Serra2015DeepDesc,Mishchuk2017LocalDescNeigh} approaches. The query local descriptors are then matched with local descriptors extracted from database images to establish tentative 2D-3D matches. These tentative matches are then geometrically verified using RANSAC~\cite{RANSAC} and the camera pose is estimated via PnP. Although these methods produce a very accurate pose estimate, the computational cost of sparse keypoint matching becomes a limitation, especially for large-scale environments. The large computational cost is addressed by image retrieval-based methods~\cite{NetVLAD,Radenovic2016GEM} restricting matching query descriptors to local descriptors extracted from top-ranked database images only. Moreover, despite the recent advancements of learned keypoint detectors and descriptors~\cite{Wang2020CAPS,Dusmanu2019CVPR,Melekhov2020Stylization,Melekhov2021hndesc,Sun2021LoFTR,Zhou2021Patch2Pix,Revaud2019R2D2,Tyszkiewicz2020DISK}, extracting discriminative local descriptors which are robust to different viewpoint and illumination changes is still an open problem.

\paragraph{Absolute camera pose regression (APR)} methods aim to alleviate the limitations of structure-based methods by using a neural network that directly regresses the camera pose of a query image~\cite{kendall2015convolutional,mapnet2018, KendallC15bay, Kendall_2017_CVPR, MelekhovYKR17, Walch_2017_ICCV, Chen2021DirectPoseNet, Chen2022DFnet} that is given as input to the network. The network is trained on database images with ground-truth poses by optimizing a weighted combination of orientation and translation L2 losses~\cite{kendall2015convolutional,MelekhovYKR17}, leveraging uncertainty~\cite{Kendall2018Uncertainty}, utilizing temporal consistency of the sequential images~\cite{Walch_2017_ICCV,radwan2018vlocnet++,valada2018deep,Xue2019LocalSupportsGlobal} or using GNNs~\cite{Xue2020GnnLocalization} and Transformers~\cite{Shavit2021TransformersLocalization}. The APR methods are scalable, fast, and memory efficient since they do not require storing a 3D model. However, their accuracy is an order of magnitude lower compared to the one obtained by structure-based localization approaches and comparable with image retrieval methods~\cite{Sattler2019}. Moreover, the APR approaches require a different network to be trained and evaluated per scene when the scenes are registered to different coordinate frames.

\paragraph{Relative camera pose regression (RPR)} methods, in contrast to APR, train a network to predict relative pose between the query image and each of the top-ranked database images~\cite{Ding_2019_ICCV, LaskarMKK17, Balntas_2018_ECCV}, obtained by image retrieval~\cite{NetVLAD,Radenovic2016GEM}. The camera location is then obtained via triangulation from two relative translation estimations verified by RANSAC. This leads to better generalization performance without using scene-specific training. However, the RPR methods suffer from low localization accuracy similarly to APR.

\paragraph{Scene coordinate regression} methods learn the first stage of the pipeline in the structure-based approaches. Namely, either a random forest~\cite{BrachmannMKYGR16,cavallari2019real,CavallariGLVST17,Guzman-RiveraKGSSFI14,Massiceti2017,meng2017backtracking,meng2018exploiting,SCoRF,ValentinNSFIT15} or a neural network~\cite{Brachmann_2017_CVPR,Brachmann_2018_CVPR,Brachmann2019SampleConsensus,Brachmann_2019_ICCV_NG,brachmann2020visual,Budvytis2019,bui2018scene,Cavallari_corr_19,Li2018,Li_Ylioinas_Verbeek_Kannala_2018,Massiceti2017} is trained to directly predict 3D scene coordinates for the pixels and thus the 2D-3D correspondences are established. These methods do not explicitly rely on feature detection, description, and matching, and are able to provide correspondences densely.
They are more accurate than traditional feature-based methods at small and medium scales, but usually do not scale well to larger scenes~\cite{Brachmann_2018_CVPR,Brachmann2019SampleConsensus}. 
In order to generalize well to novel viewpoints, these methods typically rely on only local image patches to produce the scene coordinate predictions. 
However, this may introduce ambiguities due to similar local appearances, especially when the scale of the scene is large. 
To resolve local appearance ambiguities,  we introduce element-wise conditioning layers to modulate the intermediate feature maps of the network using coarse discrete location information.
We show this leads to better localization performance, and we can robustly scale to larger environments. 

\paragraph{Joint classification-regression} frameworks have been proven effective in solving various vision tasks. 
For example, \cite{rogez17cvpr,rogez19pami} proposed a classification-regression approach for human pose estimation from single images. %
In~\cite{BrachmannMKYGR16},  a joint classification-regression forest is trained to predict scene identifiers and scene coordinates. In~\cite{Weinzaepfel_2019_CVPR}, a CNN is used to detect and segment a predefined set of planar Objects-of-Interest (OOIs), and then, to regress dense matches to their reference images. %
In~\cite{Budvytis2019}, scene coordinate regression is formulated as two separate tasks of object instance recognition and local coordinate regression. In~\cite{Brachmann2019SampleConsensus}, multiple scene coordinate regression networks are trained as a mixture of experts along with a  gating network which assesses the relevance of each expert for a given input, and the final pose estimate is obtained using a novel RANSAC framework, \ie, Expert Sample Consensus
(ESAC). 
In contrast to existing approaches, in our work, we use spatially dense discrete location labels   defined for all pixels, and propose FiLM-like~\cite{film} conditioning layers to propagate information in the hierarchy. 
We show that our novel framework allows us to achieve high localization accuracy with one single compact model.

\paragraph{Transformers} are already shown to have a positive impact on the problem of visual localization. Shavit \etal~\cite{Shavit2021TransformersLocalization} show that multi-headed transformer architectures can be used to improve end-to-end absolute camera pose localization in multiple scenes with a single trained model. Similarly, SuperGlue, LoFTR and COTR~\cite{sarlin2020superglue,Sun2021LoFTR,Jiang2021COTR} demonstrate the usefulness of transformer architectures in learning local descriptor models. Inspired by the above success, the paper proposes methods to extend transformer architecture to the structured localization method.

\section{Problem Formulation and notation}
The goal of camera pose estimation is to predict the 6-DoF pose $p(x)\in \mathbb{R}^6$ for an RGB image $x$. Handling camera pose estimation as dense 3D coordinate scene regression is performed by first predicting the corresponding 3D coordinates of a known 3D environment for each pixel of an image, given by $\hat{y}(x)$.
As a second and final step, these 2D-3D correspondences are then fed into the PnP algorithm that estimates the camera pose. 
In this work, we focus on function $f: [0,1]^{W\times H \times 3} \rightarrow \mathbb{R}^{w\times h \times 3}$, $w = W/8$ and $h = H/8$\footnote{The spatial resolution of the prediction is smaller, by a factor of 8, than that of the input image. The 3D coordinate predictions are provided for a down-sampled version of the image, which is aligned with the use of deep CNNs that inherently perform such down-sampling.}, that provides such 3D coordinate predictions given an input image $x$, \ie $\hat{y}(x) = f(x)$.
The known 3D environment is represented by a set of training images, with known ground-truth labels per pixel in the form of 3D coordinates.
The training set comprises pairs of the form $(x,y(x))$ for image $x$ and ground-truth 3D coordinates $y(x)$. 
In case ground-truth is available only sparsely, \ie on small part of the image pixels, a corresponding binary mask $m(x)\in \{0,1\}^{w\times h}$ denotes which are the valid pixels. 
The value of ground-truth or prediction at a particular pixel is denoted by $i$, \eg $y(x)_i$ for the ground-truth 3D coordinate of pixel $i$.
\begin{figure*}[t]
\begin{center}
\includegraphics[width=0.98\textwidth]{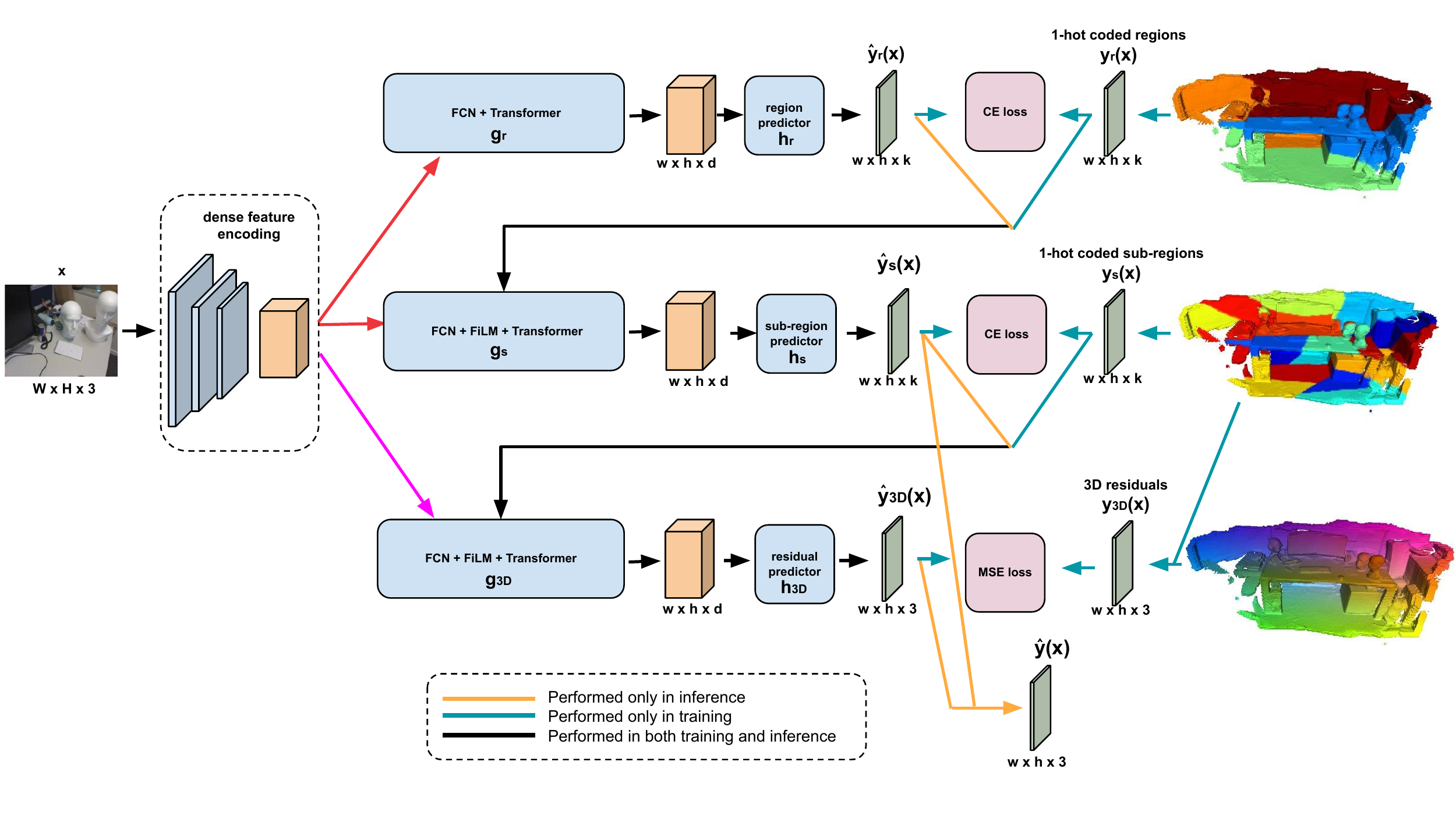}
\end{center}
\vspace{-10pt}
\caption{
\textbf{An overview of the proposed HSCNet++.}
The figure shows the network architecture of the proposed HSCNet++. 
The depicted losses correspond to the case of learning with dense ground-truth. 
Feature maps from different parts of the dense feature encoder are inputted to $g_r, g_s, g_{3D}$. This is represented by \textcolor{red}{red} and \textcolor{magenta}{magenta} arrows in an abstract way, while the detailed architecture is presented in Figure~\ref{fig:arch_details}.
}
\label{fig:arch}
\end{figure*}

\section{HSCNet++ with Dense Supervision}
\label{sec:HSCNet}
\subsection{Hierarchical Scene Coordinate Prediction}
\paragraph{HSCNet.}
A baseline \emph{conventional} approach for this task is to use a fully convolutional network (FCN) that maps input images to 3D coordinate predictions and is trained with a regression loss. 
The proposed HSCNet extends this scheme by constructing a hierarchy of  labels, from coarse-level to fine-level, and by adding extra layers to predict those labels.
Hierarchical discrete labels are defined by partitioning the ground-truth 3D points of the scene with hierarchical k-means. The number of levels in the hierarchy is fixed to 2 in this work.
In this way, in addition to the ground-truth 3D scene coordinates, each  pixel in a training image is also associated with two labels, namely region and sub-region labels, obtained at different levels of the clustering hierarchy. 
Region and sub-region labels are denoted by one-hot encodings $y_r(x)\in \{0, 1\}^{w\times h\times k}$ and $y_s(x)\in \{0,1\}^{w\times h\times k}$, respectively.
The fine-level information is given by the residual between the ground-truth 3D point and the corresponding sub-region center, which we denote by $y_{3D}(x)\in \mathbb{R}^{w\times h\times 3}$. 
Ground-truth 3D pixel coordinates $y(x)$ are replaced by $y_r(x)$, $y_s(x)$, and $y_{3D}(x)$.
Sub-region centers and residuals, when combined by addition, compose the pixel 3D coordinates.

We add two classification branches for regions and sub-regions, which provide the label predictions in the form of the k-dimensional probability distributions, and a regression branch for residual prediction. Regions, sub-region and residual predictions are denoted by $\hat{y}_r(x)$, $\hat{y}_s(x)$, and $\hat{y}_{3D}(x)$, respectively. 
A key ingredient is to propagate coarse region information to inform the predictions at finer levels, which is achieved by conditioning layers before the classification/regression layer(s). 

\paragraph{FiLM Conditioning.} The FiLM-based~\cite{film} conditioning layers are used to encode the predicted (sub-)regions information into follow-up branches.
These layers rely on parameter generators $\gamma, \beta: \real^d \rightarrow \real^d$, and 
perform modulation of input features $z$ by
\begin{equation}
\phi(z,w) = \gamma(w) \odot z + \beta(w),
\label{equ:film}
\end{equation}
where $\odot$ is the Hadamard product, and functions $\gamma$, and $\beta$ consist of $1\times 1$ convolutions and are conditional parameter generators, \ie the parameters of the FiLM layer depend on one of the inputs.
Unlike the original FiLM layers which perform the same channel-wise modulation across the entire feature map, our conditioning layers perform a linear modulation per spatial position, \ie, element-wise multiplication and addition. Therefore, instead of  vectors, the output parameters $\gamma(w)$ and $\beta(w)$ are feature maps of the same dimensions as the input feature map. 

\subsection{HSCNet++}
\label{sec:hscnet++(d)}
HSCNet is extended by adding transformer blocks at each branch of the pipeline. The resulting architecture is referred to as HSCNet++. It integrates transformer encoders that enjoy the inherent and implicit region and sub-region information and do not require the use of conventional position encodings that are typically used with transformers.

\paragraph{Model architecture.}
The overall architecture of HSCNet++ is summarized in Fig~\ref{fig:arch}. We present the model as it operates during inference and then clarifies the differences between training and inference.
An FCN backbone is used for dense feature encoding and is denoted by $\mathcal{F}(x) \in \real^{w\times h \times d}$, mapping the input image to a dense feature tensor which represents the appearance of the input image. 
Prediction of region labels is performed first.
A mixed module, consisting of FCN with transformer and denoted by $g_r$, performs encoding processing of feature map $\mathcal{F}(x)$. The result is given by $\mathbf{x}_r = g_r(\mathcal{F}(x))$, which is then fed to the region predictor $h_r: \real^{w\times h \times d} \rightarrow \real^{w\times h \times k}$ comprised a $1\times1$ convolutional layer. Region prediction is provided by $\hat{y}_r(x) = h_r(\mathbf{x}_r)$.

After the region prediction, sub-region prediction is performed.
Feature map $\mathcal{F}(x)$ is now fed to a conditioning block. Processing is performed in a way that is conditioned on region predictions $\hat{y}_r(x)$, and the features are enhanced with the transformer by capturing global information to local features. This is denoted by function $g_s: \real^{w\times h \times d} \times \real^{w\times h \times k} \rightarrow \real^{w\times h \times d}$, \ie the output feature map depends both on the input feature map and on the predicted regions. 
Then, $\mathbf{x}_s = g_s(\mathcal{F}(x), \hat{y}_r(x))$ is fed into the sub-region predictor (similar to the region predictor)
$\hat{y}_s(x) = h_s(\mathbf{x}_s)$.
Conditioning on region predictions is a way to jointly encode appearance and region predictions. Therefore, conditioning on region predictions is used to improve sub-region predictions.

Now, the last part of continuous residual is performed.
Similar to the earlier stage,
feature map $f(x)$ is processed by conditioning on sub-region predictions $\hat{y}_s(x)$. This is denoted by function $g_{3D}: \real^{w\times h \times d} \times \real^{w\times h \times k} \rightarrow \real^{w\times h \times d}$.
Then, $\mathbf{x}_{3D} = g_{3D}(f(x), \hat{y}_s(x))$ is fed into the 3D residual predictor to obtain $\hat{y}_{3D}(x) = h_{3D}(\mathbf{x}_{3D})$, where $h_{3D}: \real^{w\times h \times d} \rightarrow \real^{w\times h \times 3}$ and consists of $1\times1$ convolution.

\paragraph{Conditioning with transformer (w/ Txf).}
Conditioning blocks $g_s$ and $g_{3D}$ consist of FiLM layers, parameter generators and transformers. The parameter generators consist of several 1x1 FCN layers and take predicted (sub-)regions, $\hat{y}$ from previous layers as input. The FiLM layers then condition the input features $\mathcal{F}(x)$ with the output of parameter generators outputs using Eq.~\ref{equ:film}. Therefore, appearance information, as indicated by the input feature map, and position information is jointly encoded. We add transformer encoders right after each FiLM layer in the conditioning blocks. As FiLM layers encode both appearance and position, the requirement of conventional 2D positional encoding~\cite{vaswani2017attention} is not needed. It is to be noted that the spatial information encoded by FiLM depends on the network predictions. To the best of our knowledge, such form of spatial encoding for transformers has not appeared in the computer vision or machine literature before.

The vanilla transformer has a quadratic computation cost $\mathcal{O}(n^2)$ with the length of input features, which is computationally unaffordable in our case as we adopt a semi-dense feature map ($\mathcal{F}(x) \in \mathbb{R}^{(w \times h) \times d}$ ) as input for the scene coordinate prediction. Inspired by~\cite{katharopoulos2020transformers, Sun2021LoFTR}, we apply the linear transformer~\cite{katharopoulos2020transformers} to speed up this process and keep it in the sparse ground-truth label setting. The linear transformer considers the self-attention as a linear dot-product of kernel feature maps and leverages the associativity property of matrix
products to reduce the computational complexity to $\mathcal{O}(n)$. Consequently, the additional transformer modules have a negligible impact on our running time. 

\paragraph{Training.}
When training with dense supervision,  the following losses are adopted. Classification loss $\ell_c$ is applied to the output of the two classification branches, 
\begin{equation}
    \ell_c = \ell_{ce}(\hat{y}_r(x),y_r(x) ) +   \ell_{ce}(\hat{y}_s(x),y_s(x) )
\end{equation}
Where $\ell_{ce}$ is the cross-entropy loss. Additionally, 
regression loss $\ell_r$, in particular mean squared error, is applied on $\hat{y}_{3D}(x)$ and $y_{3D}(x)$.
The total loss $\mathcal{L}$ is a weighted summation of the two classification losses and the regression loss.
\begin{equation}
    \mathcal{L} = \lambda_1\ell_c + \lambda_2\ell_r
\end{equation}
Where $\lambda_1$ and $\lambda_2$ are the weights for each term. We observe that the regression prediction is more sensitive to localization performance. Thus, a larger weight is assigned to the $\ell_r$.

\paragraph{Inference.}
During inference, the predicted 3D coordinates $\hat{y}(x)$ and their corresponding 2D pixels are fed into the PnP-RANSAC loop to estimate the 6-DoF camera pose. 
These predicted 3D coordinates are obtained by simply adding the center of predicted sub-regions $\hat{y}_s(x)$ and predicted residuals $\hat{y}_{3D}(x)$. 

We differentiate on how conditioning is performed during training and inference as shown in Fig~\ref{fig:arch}. 
At training time, conditioning is performed using the ground truth~(sub-)region labels, \ie $y_r(x)$ and $y_s(x)$ are the second inputs of the conditioning blocks. 
At test time, conditioning is performed using predicted (sub-)region labels. In particular, the one-hot encodings of the $\argmax$ operation of $\hat{y}_r(x)$ and $\hat{y}_s(x)$ are the second inputs of the conditioning blocks.

\section{HSCNet++ with Sparse Supervision}
When only sparse ground truth of 3D coordinates, indicated by mask $m(x)$ for image $x$, is available,
the straightforward approach is to apply the loss only on pixels where the mask value is 1, which we refer to as  \emph{valid pixel}. Instead, we propose to perform propagation of the available labels to nearby pixels and use two additional losses that are appropriately handling the scarcity of the labels.

\paragraph{Label propagation (LP).} 
We rely on a smoothness assumption: labels do not change much in a small pixel neighborhood.
Consequently, we propagate the labels in a local neighborhood around each pixel. 
The neighborhood is defined by a square area of size $(2z+1) \times (2z+1)$.
All neighbors of a valid pixel are marked as valid too and ground-truth maps, namely $y_r(x)$, $y_s(x)$, and $y_{3D}(x)$,  are updated by replicating the label of the original pixel to the neighboring pixels.
Then, the classification and regression losses are applied on the newly obtained valid pixels after propagation.
This is seen as some form of pseudo-labeling that increases the density of the available labels. 

\paragraph{Symmetric cross-entropy loss (SCE).} Pseudo-labels are expected to include noise. This noise will typically be larger if propagation reaches background pixels starting from a foreground-object valid pixel.
The conventional cross-entropy loss that we use as classification loss is shown to be not very robust to noise in the labels
in the work of Wang \etal~\cite{wang2019symmetric}.
Inspired by their work, we use the symmetric cross-entropy loss given by
\begin{equation}
\ell_{r-rce}(x,i) = \hat{y}_r(x)_i \log y_r(x)_i,
\label{equ:rce}
\end{equation}
compared to the conventional one defined as follows:
\begin{equation}
\ell_{r-ce}(x,i) = y_r(x)_i \log \hat{y}_r(x)_i,
\label{equ:ce}
\end{equation}
for pixel $i$ for the region prediction and similarly for the sub-region prediction too.
Computational problems are simply solved by setting $\log 0$ equal to a constant value.
The total classification loss is just a weighted summation $\ell_{sce} =\lambda_{ce} \ell_{ce} + \lambda_{rce} \ell_{rce}$

\paragraph{Re-projection error loss (Rep).}
We additionally use a re-projection error loss that does not require any labels and, therefore, does not get influenced by noise in the pseudo-labels.
Nevertheless, we do not apply this on all pixels to avoid background pixels, but rather apply it on only all valid pixels after the label propagation, therefore staying near the original valid pixels.
We use the angle-based re-projection error as a loss.
Given ground-truth camera pose $F$, the loss for pixel $i$ of image $x$, whose 2D coordinates in the image are denoted by $p_i$, is given by

\begin{equation}
\ell_{rep}(x,i) = ||\gamma_{i} F^{-1}\hat{y}_i(x) - fC^{-1}p_i||,
\label{equ:rep}
\end{equation}
where $\gamma_i = ||fC^{-1}p_i||/||F^{-1}\hat{y}(x)_i||$, $f$ is the focal length, and $C$ is the intrinsic matrix. 
Note that re-projection loss is not added to the total loss in the beginning epochs for a fast training convergence. 
Similar to our dense setting, The total loss for sparse supervision is the weighted summation of regression loss, symmetric classification loss, and re-projection loss,
$\ell_{sparse} = \ell_{sce} + \lambda_2 \ell_r + \lambda_3 \ell_{rep}$.

\section{Experiments}

In this section, we discuss the experimental setup and employed datasets, present our results, and compare our approach to state-of-the-art localization methods.

\begin{figure*}[t!]
    \centering

    \begin{tabular}{c}
    \hspace{25pt}
        \includegraphics[width=0.95\textwidth]{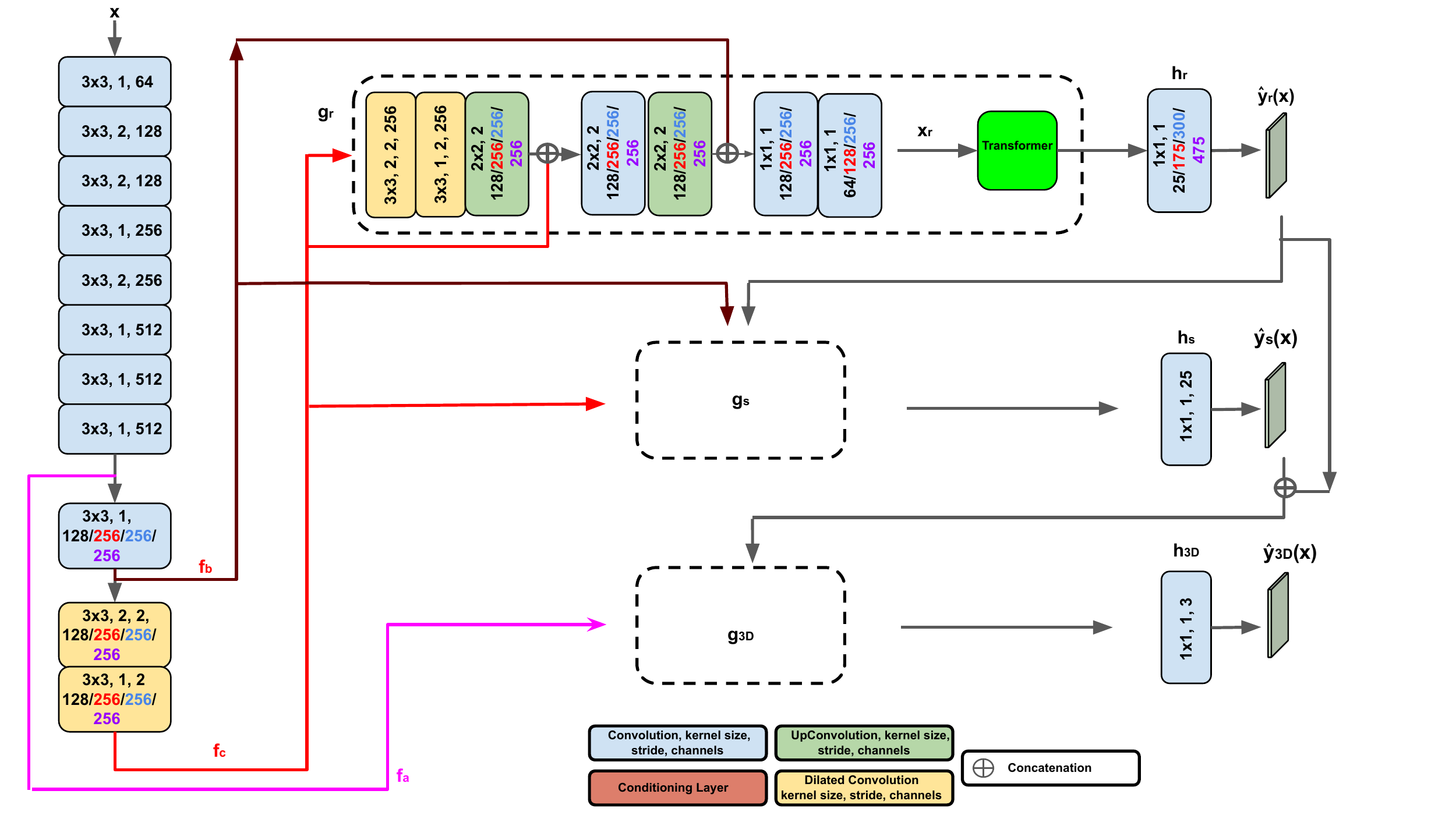} \\
        \\
        (a) Main Network \\
    \end{tabular}

    \begin{tabular}{c}
        \includegraphics[width=0.95\textwidth]{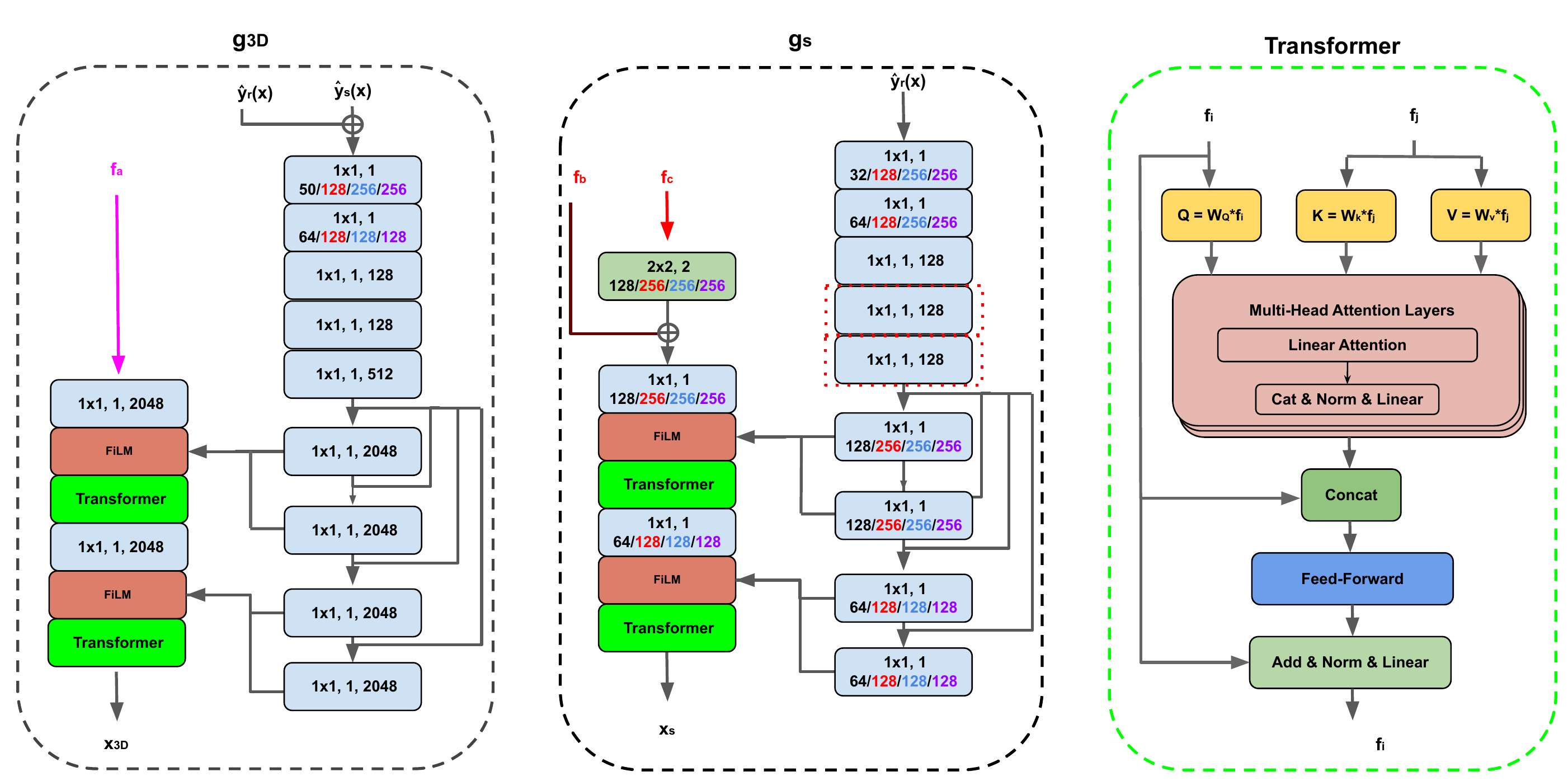} \\
        \\
        (b) FiLM Conditioning Network \\
    \end{tabular}
\caption{
\textbf{HSCNet++ detailed architecture.}
The figure shows the detailed network architecture of the main pipeline and the FiLM conditioning network. For experiments on the combined scenes we added two more layers in the first conditioning generator, $g_s$ that are marked in (dotted) red. We also roughly doubled the channel counts that are highlighted in \textcolor{red}{red}, \textcolor{cyan}{cyan} and \textcolor{violet}{violet} for i7-Scenes, i12-Scenes and i19-Scenes, respectively.
}
\label{fig:arch_details}
\end{figure*}

\begin{table*}[t!]
        \centering
        \fontsize{7}{8}\selectfont
        \renewcommand{\tabcolsep}{4pt}
		\begin{tabular}{l | c c | c c | c c | c c | c c | c c | c c | c }
		\toprule
		\multirow{3}{*}{Method} & \multicolumn{14}{c|}{7-Scenes} & \multirow{3}{*}{Accuracy} \\
		 & \multicolumn{2}{c}{Chess} & \multicolumn{2}{c}{Fire} & \multicolumn{2}{c}{Heads} & \multicolumn{2}{c}{Office} & \multicolumn{2}{c}{Pumpkin} & \multicolumn{2}{c}{Red Kitchen} & \multicolumn{2}{c|}{Stairs} & \\
		 & $\bm{t}$, cm & $\bm{r}$, $^\circ$ & $\bm{t}$, cm & $\bm{r}$, $^\circ$ & $\bm{t}$, cm & $\bm{r}$, $^\circ$ & $\bm{t}$, cm & $\bm{r}$, $^\circ$ & $\bm{t}$, cm & $\bm{r}$, $^\circ$ & $\bm{t}$, cm & $\bm{r}$, $^\circ$ & $\bm{t}$, cm & $\bm{r}$, $^\circ$ & \\
	     \midrule
          MapNet~\cite{mapnet2018} & 8.0 & 3.30 & 27.0 & 11.70 & 18.0 & 13.30 & 17.0 & 5.20 & 22.0 & 4.00 & 23.0 & 4.90 & 30.0 & 12.10&-- \\
          Geometric PoseNet~\cite{Kendall_2017_CVPR} & 13.0 & 4.50 & 27.0 & 11.30 & 17.0 & 13.00 & 19.0 & 5.60 & 26.0 & 4.80 & 23.0 & 5.40 & 35.0 & 12.40&--  \\
          AttTxf~\cite{Shavit2021TransformersLocalization} & 11.0 & 4.66 & 24.0 & 9.60 & 14.0 & 12.19 & 17.0 & 5.66 & 18.0 & 4.44 & 17.0 & 5.94 & 26.0 & 8.45&-- \\
          LSTM-Pose~\cite{Walch_2017_ICCV} & 24.0 & 5.80 & 34.0 & 11.90 & 21.0 & 13.70 & 30.0 & 8.10 & 33.0 & 7.00 & 37.0 & 8.80 & 40.0 & 13.70 &-- \\
          AnchorNet\cite{saha2018improved} & 6.0 & 3.90 & 16.0 & 11.10 & 9.0 & 11.20 & 11.0 & 5.40 & 14.0 & 3.60 & 13.0 & 5.30 & 21.0 & 11.90 &--\\
	     LENS~\cite{moreau2021lens} & 3.0 & 1.30 & 10.0 & 3.70 & 7.0 & 5.80 & 7.0 & 1.90 & 8.0 & 2.20 & 9.0 & 2.20 & 14.0 & 3.60 & -- \\
          AS~\cite{sattler2016efficient} & 3.0 & 0.87 & \underline{2.0} & 1.01 & \textbf{1.0} & 0.82 & 4.0 & 1.15 & 7.0 & 1.69 & 5.0 & 1.72 & 4.0 & 1.01 & 68.7 \\
          HLoc~\cite{sarlin2019coarse} & \underline{2.0} & 0.85 & \underline{2.0} & 0.94 & \textbf{1.0} & \underline{0.75} & 3.0 & 0.92 & 5.0 & 1.30 & 4.0 & 1.40 & 5.0 & 1.47 & 73.1 \\
	     PixLoc~\cite{sarlin2021back} & \underline{2.0} & 0.80 & \underline{2.0} & \textbf{0.73} & \textbf{1.0} & 0.82 & 3.0 & 0.82 & 4.0 & 1.21 & \textbf{3.0} & 1.20 & 5.0 & 1.30 & 75.7 \\
	     VS-Net~\cite{huang2021vs} & \textbf{1.5} & \textbf{0.50} & \textbf{1.9} & 0.80 & 1.2 & \textbf{0.70} & \underline{2.1} & \textbf{0.60} & 3.7 & 1.00 & 3.6 & 1.10 & \textbf{2.8} & \textbf{0.80} & -- \\
	     SFT-CR~\cite{9437699} & 2.1 & 0.70 & \underline{2.0} & \underline{0.78} & \underline{1.1} & 0.81 & 2.4 & 0.66 & \underline{3.4} & \underline{0.98} & \underline{3.4} & \textbf{1.06} & 3.5 & 0.97 & -- \\
          DSAC++~\cite{Brachmann_2018_CVPR} & \underline{2.0} & \textbf{0.50} & \underline{2.0} & 0.90 & \textbf{1.0} & 0.80 & 3.0 & 0.70 & 4.0 & 1.10 & 4.0 & 1.10 & 9.0 & 2.60 & 76.1 \\
          DSAC$^\star$(3D)~\cite{brachmann2020visual} & \underline{2.0} & 1.10 & \underline{2.0} & 1.24 & \textbf{1.0} & 1.82 & 3.0 & 1.15 & 4.0 & 1.34 & 4.0 & 1.68 & \underline{3.0} & 1.16 & \underline{85.2} \\
	     \midrule
          Reg-only~\cite{li2020hscnet} & \underline{2.0} & 0.70 & \underline{2.0} & 0.90 & \textbf{1.0} & 0.80 & 3.0 & 0.90 & 4.0 & 1.10 & 5.0 & 1.40 & 4.0 & 1.00 & 74.7 \\
	     HSCNet~\cite{li2020hscnet} & \underline{2.0} & 0.70 & \underline{2.0} & 0.90 & \textbf{1.0} & 0.90 & 3.0 & 0.80 & 4.0 & 1.00 & 4.0 & 1.20 & \underline{3.0} & \textbf{0.80} & 84.8 \\
         HSCNet++ & \underline{2.0} & \underline{0.63} & \underline{2.0} & 0.79 & \textbf{1.0} & 0.80 & \textbf{2.0} & \underline{0.65} & \textbf{3.0} & \textbf{0.85} & \textbf{3.0} & \underline{1.09} & \underline{3.0} & \underline{0.83} & \textbf{88.7} \\
         \bottomrule
		\end{tabular}
	\caption{\textbf{Indoor localization: individual scene setting (7-Scenes)}. For each scene of 7-Scenes dataset we report the median translation ($\bm{t}$, cm) and orientation ($\bm{r}$, $^\circ$) error. The best and second best results are in \textbf{bold} and \underline{underlined}. Note that except VS-Net~\cite{huang2021vs} and SFT-CR~\cite{9437699}, the rest results are reported in centimeter precision for translation error.
	\label{tbl:7S_individual}
	}

\end{table*}

\begin{table}[t!]
	\centering
        \fontsize{5}{6}\selectfont
        \renewcommand{\tabcolsep}{1pt}
		\begin{tabular}{l | c c c|  c c c|  c c c| c c c   }
		\toprule
		 \multirowcell{3}{Scenes} & \multicolumn{12}{c}{Methods} \\
		 & \multicolumn{3}{c}{Reg-only\cite{li2020hscnet}} & \multicolumn{3}{c}{DSAC*(3D)\cite{brachmann2020visual}} & \multicolumn{3}{c}{HSCNet~\cite{li2020hscnet}} & \multicolumn{3}{c}{HSCNet++} \\
		 & $\bm{t}$, cm & $\bm{r}$, $^\circ$ & Acc & $\bm{t}$, cm & $\bm{r}$, $^\circ$ & Acc & $\bm{t}$, cm & $\bm{r}$, $^\circ$ & Acc& $\bm{t}$, cm & $\bm{r}$, $^\circ$ & Acc \\
	     \midrule
     Kitchen-1 & 0.8 & \textbf{0.4} &\textbf{100} & \multicolumn{2}{c}{-}&- & 0.8 & \textbf{0.4} &\textbf{100}  & \textbf{0.7} & \textbf{0.4} &\textbf{100}  \\
	     Living-1  & 1.1 & \textbf{0.4} &\textbf{100}  & \multicolumn{2}{c}{-}&- & 1.1 & \textbf{0.4} &\textbf{100}  & \textbf{1.0} & \textbf{0.4} &\textbf{100}  \\
	     Bed       & 1.3& 0.6 &\textbf{100}  & \multicolumn{2}{c}{-}&- & \textbf{0.9} & \textbf{0.4} &\textbf{100}  & 1.0 & \textbf{0.4} &\textbf{100}   \\
	     Kitchen-2 & 0.8 & 0.4&\textbf{100}  & \multicolumn{2}{c}{-}&- & \textbf{0.7} & \textbf{0.3} &\textbf{100}  & 0.8 & 0.4 &\textbf{100}   \\
	     Living-2  & 1.4 & 0.6 &\textbf{100}  & \multicolumn{2}{c}{-}&- & \textbf{1.0} & \textbf{0.4} &\textbf{100}  & \textbf{1.0} & \textbf{0.4} &\textbf{100}  \\
	     Luke      & 2.0 & 0.9 &93.8  & \multicolumn{2}{c}{-}&- & \textbf{1.2} & \textbf{0.5} &96.3  & 1.3 & 0.6 &\textbf{98.1}   \\
	     Gate362   & 1.1 & 0.5 &\textbf{100}  & \multicolumn{2}{c}{-}&- & \textbf{1.0} & \textbf{0.4} &\textbf{100}  & \textbf{1.0} & 0.5 &\textbf{100}   \\
	     Gate381   & 1.6 & 0.7 & 98.8  & \multicolumn{2}{c}{-}&- & 1.2 & 0.6 &\textbf{99.1}  & \textbf{1.1} & \textbf{0.5} &98.6  \\
	     Lounge    & 1.5 & 0.5 & 99.4 & \multicolumn{2}{c}{-}&- & 1.4 & 0.5 &\textbf{100}  & \textbf{1.3} & \textbf{0.4} &\textbf{100}  \\
	     Manolis   & 1.4&0.7 &97.2   & \multicolumn{2}{c}{-}&- & \textbf{1.1} & \textbf{0.5} &\textbf{100}  & 1.2 & \textbf{0.5} &\textbf{100}  \\
	     Floor. 5a & 1.6 & 0.7 & 97.0   & \multicolumn{2}{c}{-}&- & \textbf{1.2} & \textbf{0.5} &\textbf{98.8}  & 1.3 & \textbf{0.5} &96  \\
	     Floor. 5b  & 1.9 &0.6&93.3   & \multicolumn{2}{c}{-}&- & 1.5 & 0.5 &97.3  & \textbf{1.4} & \textbf{0.4} & \textbf{99.5} \\
         \midrule
         Accuracy & \multicolumn{3}{c}{96.4}  & \multicolumn{3}{c}{99.1}&  \multicolumn{3}{c}{99.1} & \multicolumn{3}{c}{\textbf{99.4}}  \\
         \bottomrule
		\end{tabular}

	\caption{\textbf{Indoor localization: individual scene setting (12-Scenes)}. Similar to the 7-Scenes localization benchmark, we provide the median translation ($\bm{t}$, cm), orientation ($\bm{r}$, $^\circ$) error, and accuracy with the error threshold of $5cm$ and $5^\circ$. The best accuracy results are in \textbf{bold}. 
	\label{tbl:12S}
	}

\end{table}

\subsection{Experimental setup}

\noindent\textbf{Datasets.}
We use three standard benchmarks for the evaluation; namely, 7-Scenes ~\cite{SCoRF}, 12-Scenes~\cite{valentin2016learning}, and Cambridge Landmarks~\cite{kendall2015convolutional}. 
The  7-Scenes dataset covers a volume of $\sim 6 m^3$ for each individual scene. The 3D models and ground truth poses are included in the dataset. 12-Scenes is another indoor RGB-D dataset that contains 4 large scenes with a total of 12 rooms, the volume ranges $14 \text{--} 79 m^3$ for each room.
The union of these two datasets forms the 19-Scenes dataset. 
Cambridge Landmarks dataset is a standard benchmark for evaluating scene 
coordinate methods in outdoor scenes. It is a small-scale outdoor dataset consisting of 6 individual scenes, and the ground truth pose is provided by structure-from-motion.   

Following prior work~\cite{Brachmann2019SampleConsensus}, we conduct experiments per scene, \ie the individual scenes setting, but also by training a single model on all scenes of a corresponding dataset, \ie the combined scenes setting. The combined settings of the given indoor localization benchmarks are denoted by i7-Scenes, i12-Scenes, and i19-Scenes, respectively. 

\noindent\textbf{Competing methods.} 
In this work, we compare the proposed approach with the following methods: \emph{(1)} pose regression methods that directly regress absolute or relative camera pose parameters: MapNet~\cite{mapnet2018}, Geometric PoseNet~\cite{Kendall_2017_CVPR}, AttTxf~\cite{Shavit2021TransformersLocalization}, LSTM-Pose~\cite{Walch_2017_ICCV}, AnchorNet~\cite{saha2018improved} and {LENS}~\cite{moreau2021lens}; \emph{(2)} local feature based pipelines based on SIFT such as Active Search (AS) \cite{sattler2016efficient} and Hloc~\cite{sarlin2019coarse} based on CNN descriptors; \emph{(3)}~{DSAC$^\star$(3D)}~\cite{brachmann2020visual}: the latest scene coordinate regression approach with 3D model; \emph{(4)}~{VS-Net}~\cite{huang2021vs}: scene-specific segmentation and voting; \emph{(5)}~{PixLoc}~\cite{sarlin2021back}: scene-agnostic network; \emph{(6)}~{SFT-CR}~\cite{9437699}: scene coordinate regression with global context-guidance.  In addition, we
also compare with \emph{(7)}~{ESAC}~\cite{Brachmann2019SampleConsensus} on the combined scenes. 
We also consider a baseline called \textit{Reg-only} without the hierarchical classification layers.

\begin{table}[t!]
	\centering
        \fontsize{7}{8}\selectfont
        \renewcommand{\tabcolsep}{12pt}
		\begin{tabularx}{.48\textwidth}{@{} l | S[table-format=2.1,detect-weight,detect-shape,detect-mode] S[table-format=2.1,detect-weight,detect-shape,detect-mode] S[table-format=2.1,detect-weight,detect-shape,detect-mode]}
        \toprule
        \multirow{2}{*}{Method} & \multicolumn{3}{c}{Localization Accuracy (\%)}\\
        & {i7-Scenes} & {i12-Scenes} & {i19-Scenes}  \\
        \midrule
        Reg-only~\cite{li2020hscnet}  & 37.9 & 5.0 & 5.7                       \\

        ESAC~\cite{Brachmann2019SampleConsensus} & 70.3 & 97.1 & 88.1            \\
        \midrule
        HSCNet~\cite{li2020hscnet}  & 83.3 & 99.3 &  92.5 \\
        HSCNet++    & \bfseries 88.3 & \bfseries 99.5 & \bfseries 93.6 \\
        \bottomrule
        \end{tabularx}
	\caption{\textbf{Indoor localization: combined scene setting}. The table presents average localization accuracy under $5cm/5^\circ$ of baseline models and proposed methods on i7-Scenes, i12-Scenes, and i19-Scenes datasets.}
	\label{tbl:i_summary}
\end{table}

\noindent\textbf{Evaluation metrics.}
We report the median translation and orientation error ($cm$,$^\circ$) as well as the accuracy of test images under the threshold of ($5cm, 5^\circ$) on indoor scenes. 
On Outdoor Cambridge Landmarks~\cite{kendall2015convolutional}, we report only the median pose error as in previous methods~\cite{brachmann2020visual, Brachmann_2017_CVPR, li2020hscnet}. 

\noindent\textbf{Training details.} 
We generate the region labels by hierarchical K-means. For 7-Scenes, 12-Scenes, and Cambridge landmarks, we adopt 2-level ground truth labels with a branching factor of $25$ for all the levels. Furthermore, for combined scenes, i7-Scenes, i12-Scenes, and i19-Scenes, the first level branching factor is set to $7\times25$, $12\times25$, and $19\times25$, respectively. 
For the individual scene setting, training is performed for 300K iterations with Adam optimizer. For the combined scenes,
the number of iterations is set to 900K. Throughout all experiments, we use a batch size of 1 with the initial learning rate of $10^{-4}$. 

The classification loss weights $\lambda_1$ is set to 1 for all datasets, while regression loss weight $\lambda_2$ is 10 for single scenes and 100000 for combined scenes. In the sparse supervision setting, $\lambda_{ce}$ and $\lambda_{rce}$ are set to 0.1 and 1, respectively, while $\lambda_2$ follows the dense setting, and $\lambda_3$ is increased from 0 to 0.1 after first 10 epochs. We initialize the network by training with $l_r$ using pseudo-label coordinates and later also add $l_{rep}$ after 10 epochs. 
When training with sparse supervision, we select the neighborhood size $z = 5$ to propagate labels, and use the cluster centers obtained from dense scene coordinates for a direct comparison.

Data augmentation is also effective in increasing the prediction accuracy. Thus,  similar to HSCNet~\cite{li2020hscnet}, we randomly augment training images using translation, rotation, scaling and shearing by uniform sampling from [-20\%, 20\%], [-30$^\circ$, 30$^\circ$], [0.7, 1.5], [-10$^\circ$, 10$^\circ$] respectively. In addition, images are augmented with additive brightness uniformly sampled from [-20, 20].

\noindent\textbf{Pose estimation.}  We follow the same PnP-RANSAC pipeline and parameters setting as in~\cite{Brachmann_2018_CVPR}. The inlier threshold and the softness factor are set to $\tau = 10$ and $\beta = 0.5$, respectively. We randomly select 4 correspondences to formulate a minimal set for a PnP algorithm to generate a camera pose hypothesis, and a set of 256 initial hypotheses are sampled. Similar to ~\cite{Brachmann_2018_CVPR, brachmann2020visual}, a pose refinement process is performed until convergence for a maximum of 100 iterations.

\noindent\textbf{Architecture details.}  
The detailed architecture of HSCNet++ is shown in Fig~\ref{fig:arch_details}; we also visualize the block details of the FiLM conditioning network and the transformer modules. By removing the transformer layers, we derive the architecture of HSCNet. Additionally, the number of channels in the last branch, $g_{3D}$ of HSCNet is 4096, while it is 2048 for HSCNet++ that reduces memory cost (c.f. Sec.~\ref{sec:eff}). For experiments on the combined scenes we added two more layers in the first conditioning generator, $g_s$ that are marked in (dotted) red. We also roughly doubled the channel counts that are highlighted in red, cyan and violet for i7-Scenes, i12-Scenes and i19-Scenes, respectively. For individual scenes, we add 2 multi-head attention layers (MHA) to both classification and regression conditioning blocks, while in the combined setting, the number of MHA is set to 5. 
\begin{table*}[t!]
        \centering
        \fontsize{7}{8}\selectfont
        \renewcommand{\tabcolsep}{5pt}
		\begin{tabular}{l | c c | c c | c c | c c | c c }
		\toprule
		\multirow{3}{*}{Method} & \multicolumn{10}{c}{Cambridge}  \\
		 & \multicolumn{2}{c}{Kings College} & \multicolumn{2}{c}{Great Court} & \multicolumn{2}{c}{Old Hospital} & \multicolumn{2}{c}{Shop Facade} & \multicolumn{2}{c}{St Mary Church}  \\
		 & $\bm{t}$, cm & $\bm{r}$, $^\circ$ & $\bm{t}$, cm & $\bm{r}$, $^\circ$ & $\bm{t}$, cm & $\bm{r}$, $^\circ$ & $\bm{t}$, cm & $\bm{r}$, $^\circ$ & $\bm{t}$, cm & $\bm{r}$, $^\circ$ \\
	     \midrule
          AS~\cite{sattler2016efficient} & 24&0.13 & 13&0.22 &20&0.36 &\textbf{4}&0.21& 8&0.25\\
          HLoc~\cite{sarlin2019coarse} & 16&\textbf{0.11} &\textbf{12}&0.20& \textbf{15}&\textbf{0.30} &\textbf{4}&\textbf{0.20}& \textbf{7}&\textbf{0.21}\\
	     PixLoc~\cite{sarlin2021back} & 14 & 0.24 & 30 & 0.14 & 16 & 0.32 & 5 & 0.23 & 10 & 0.34 \\
	     VS-Net~\cite{huang2021vs} & 16 & 0.20 & 22 & \textbf{0.10} & 16 & \textbf{0.30} & 6 & 0.30 & 8 & 0.30 \\
          DSAC++~\cite{Brachmann_2018_CVPR} & \textbf{13} & 0.40 & 40 & 0.20 & 20 & {0.30} & 6 & 0.30 & 13 & 0.40 \\
          DSAC$^\star$(3D)~\cite{brachmann2020visual} & 15 & 0.30 & 49 & 0.30 & 21 & 0.40 & 5 & 0.30 & 13 & 0.40 \\
	     \midrule
	     HSCNet~\cite{li2020hscnet} & 18 & 0.30 & 28 & 0.20 & 19 & \textbf{0.30} & 6 & 0.30 & 9 & 0.30 \\
         HSCNet++ & 19 & 0.34 & 39 & 0.23 & 20 & 0.31 & 6 & 0.24 & 9 & 0.27 \\
         \bottomrule
		\end{tabular}
	\caption{\textbf{Outdoor localization: individual scene setting (Cambridge)}. For each scene of the dataset we report the median translation ($\bm{t}$, cm) and orientation ($\bm{r}$, $^\circ$) error. The best results are in \textbf{bold}. 
	\label{tbl:cam_dense}
	}

\end{table*}

\subsection{Results for HSCNet and HSCNet++}
\noindent\textbf{Individual scenes setting.} 
We present results on 7-Scenes and 12-Scenes in Table~\ref{tbl:7S_individual} and Table~\ref{tbl:12S}, accordingly. All models are trained and evaluated individually on each scene of the corresponding dataset. Results show that HSCNet is still competitive with respect to methods published later. With the addition of transformers, HSCNet++
further boosts the average performance by ~4\% on 7-Scenes and obtains the best accuracy on 7-Scenes among the competitors. 

\noindent\textbf{Combined scenes setting.} 
To test the scalability of scene-coordinate regression methods, we go beyond small-scale environments such as individual scenes in 7-Scenes and 12-Scenes and use the combined scenes, \ie i7-Scenes, i12-Scenes, and i19-Scenes by combining the former datasets. 

Results on the combined scenes setting  presented in Table \ref{tbl:i_summary} including comparison with the regression-only baseline and ESAC. Results show that our method scales well with increase in number of scenes compared to \textit{Reg-only} baseline. 
It is to be noted that ESAC requires training and storing multiple networks specializing in local parts of the environment, whereas our approach requires only a single model. Results show that our approach outperforms ESAC on i7-Scenes and i12-Scenes, while performing comparably on i19-Scenes (87.9\% \vs 88.1\%). ESAC and our approach could be combined for very large-scale scenes, but we do not explore this option in this work. HSCNet++ advances the state-of-the-art on all datasets, demonstrating the utility of transformers for this task. 

\noindent{\textbf{Cambridge Landmarks}.} Table~\ref{tbl:cam_dense} reports the results of three types of visual localization methods on Cambridge landmarks. AS~\cite{sattler2016efficient} and Hloc~\cite{sarlin2019coarse} estimate the camera poses with sparse SfM ground truth. DSAC++, DSAC* and our approaches train a scene-coordinate regression model with MVS-densified depth maps, VS-Net leverage the hybrid of the two. Both HSCNet and HSCNet++ perform better than other scene coordinate methods DSAC++ and DSAC*. The performance is comparable to more recent approaches. However, we observe that the models trained with MVS-densified pseudo ground truth show a lightly worse performance compared to the approaches that use the sparse SfM 3D map. HSCNet++ shows even worse performance by adding the transformer modules. Such results motivated us to extend the HSCNet++ to train with sparse supervision and our hypothesis is that the MVS densification introduces more noise to the dense supervision. The HSCNet++(S) performance on Cambridge landmarks in Sec.~\ref{sec:hscnet++} verified our hypothesis. 

\subsection{Ablations: HSCNet} \label{sec:ablation}

\noindent{\textbf{Data augmentation.} 
Using geometric and color data augmentation provides robustness to lighting and viewpoint changes~\cite{DeTone_2018_CVPR_Workshops,Melekhov2021hndesc}. 
We investigate the impact of data augmentation and summarize the obtained results in Table~\ref{tab:hscnet_ablation_augmentation_conditioning}. 
Applying data augmentations leads to better localization accuracy.
Note that without data augmentation, the proposed approach  still provides comparable results to state of the art methods (\cf ESAC~\cite{Brachmann2019SampleConsensus} in Table~\ref{tbl:i_summary} \vs row 3 of Table~\ref{tab:hscnet_ablation_augmentation_conditioning}). 

\noindent{\textbf{Conditioning  mechanism.}
The two key components of HSCNet are the coarse-to-fine joint classification-regression module and its combination with the conditioning mechanism. Their impact is evaluated and results are shown in Table~\ref{tab:hscnet_ablation_augmentation_conditioning}.
We train a variant of our network without the conditioning  mechanism, \ie we remove all the conditioning generators and layers.
The network still estimates scene coordinates in a coarse-to-fine manner by using the predicted location labels, but there is no coarse location information that is fed to influence the network activations at the finer levels. 
Results indicate the importance of the conditioning mechanism for accurate scene coordinate prediction.
The regression only baseline fails to achieve good performance, which pronounces the benefit of the proposed hierarchical scheme.

\begin{table}[t!]
    \begin{subtable}[t!]{\linewidth}
    \centering
    \fontsize{7}{8}\selectfont
    \begin{tabular}{l|ccc}
    \toprule
    \multirow{2}{*}{Method} & \multicolumn{3}{c}{Localization Accuracy (\%)}\\
      & i7-Scenes & i12-Scenes & i19-Scenes \\
    \midrule
    HSCNet~\cite{li2020hscnet} & 83.3 & 99.3 &  92.5 \\
    \hspace{5pt} w/o conditioning  & 70.3 & 97.1 & 88.1 \\
    \hspace{5pt} w/o augmentation   & 71.5 & 98.7 & 87.9 \\
    \bottomrule
    \end{tabular}
    \caption{Data augmentation and conditioning mechanism}\label{tab:hscnet_ablation_augmentation_conditioning}
    \end{subtable}
    \vspace{5pt}

    \begin{subtable}[t!]{0.47\linewidth}
    \centering
    \fontsize{6}{7}\selectfont
    \renewcommand{\tabcolsep}{3pt}
    \begin{tabular}{l|c}
        \toprule
        Label hierarchy & Accuracy, \% \\
        \midrule
        9$\times$9 & 82.9 \\
        49$\times$49 & 85.0 \\
        10$\times$100$\times$100 & 85.9 \\
        10$\times$100$\times$100$\times$100 & 85.5 \\
        625 & 85.3 \\
        25$\times$25 & 84.8 \\
        \bottomrule
    \end{tabular}
    \caption{Label hierarchy: 7-Scenes dataset}
    \end{subtable}%
    \hfill
    \begin{subtable}[t!]{0.47\linewidth}
    \centering
    \fontsize{6}{7}\selectfont
    \renewcommand{\tabcolsep}{3pt}
    \begin{tabular}{l|c}
        \toprule
        Label hierarchy & Accuracy, \% \\
        \midrule
        63$\times$9 & 80.6 \\
        343$\times$49 & 83.7 \\
        70$\times$100$\times$100 & 83.0 \\
        70$\times$100$\times$100$\times$100 & 82.1 \\
        7$\times$25$\times$25 & 83.0 \\
        175$\times$25 & 83.3 \\
        \bottomrule
    \end{tabular}
    \caption{Label hierarchy: i7-Scenes dataset}
    \end{subtable}
\caption{\textbf{Ablation for HSCNet}. Average pose accuracy obtained with different hierarchy settings. 
  The models with  4-level label hierarchy are classification-only, \ie the final  regression layer is omitted}
  \label{tab:eval:ll}
\end{table}

\begin{table*}[t!]

    \begin{subtable}[t]{0.99\textwidth}
            \centering \fontsize{8}{9}\selectfont
	    \renewcommand{\tabcolsep}{10pt}
		\begin{tabular}{l|c|ccccccc|c}
		\toprule 
		{Method} & {\#MHA} & Chess & Fire & Heads & Office & Pumpkin & Kitchen & Stairs & Average \\
		\midrule
		HSCNet++$^\dagger$ &5 & 95.3 & 96.0 & 98.4 & 88.6 & 63.7 & 71.8 & 80.4 & 84.9 \\
		HSCNet++$^\dagger$ & 8& 94.9 & 95.3 & 98.4 & 88.3 & 63.7 & 70.0 & 79.5 & 84.3 \\
		\midrule
		HSCNet++ &5  & \textbf{98.2} & \textbf{96.6} & \textbf{99.6} & \textbf{90.8} & \textbf{72.1} & \textbf{76.8} & \textbf{83.7} & \textbf{88.3} \\
		\bottomrule
        \end{tabular}
	\caption{The impact of increasing the number of MHA layers \#MHA. Without intermediate transformers at the classification branches (only $t_{3D}$ is used), adding additional \#MHA layers to HSCNet++$^\dagger$ does not improve performance.}\label{tbl:ablation_number_of_params}
    \end{subtable}
    
    \vspace{10pt}

    \begin{subtable}[t]{0.99\textwidth}
        \centering \fontsize{8}{9}\selectfont
	    \renewcommand{\tabcolsep}{10pt}
		\begin{tabular}{l|c|ccccccc|c}
		\toprule 
		{Method} & {Encoding} & Chess & Fire & Heads & Office & Pumpkin & Kitchen & Stairs & Average \\
		\midrule
		HSCNet++ & w/ PE & 87.4 & 60.1 & 80.3 & 79.0 & 66.9 & 67.4 & 17.7 & 65.5  \\
		HSCNet++ &  & \textbf{98.2} & \textbf{96.6} & \textbf{99.6} & \textbf{90.8} & \textbf{72.1} & \textbf{76.8} & \textbf{83.7} & \textbf{88.3} \\
		\bottomrule
        \end{tabular}
	\caption{Positional encoding. PE: conventional positional encoding with sine and cosine functions.}\label{tbl:ablation_positional_encoding}
    \end{subtable}
    
    \vspace{10pt}
    
    \begin{subtable}[t]{0.99\textwidth}
    \centering \fontsize{8}{9}\selectfont
	    \renewcommand{\tabcolsep}{12.8pt}
    \begin{tabular}{l|ccccccc|c}
    \toprule
    {Method} & Chess & Fire & Heads & Office & Pumpkin & Kitchen & Stairs & Average \\
    \midrule
    HSCNet & 75.1 & 65.6 & 63.0 & 77.2 & 66.1 & 72.4 & 51.5 & 67.3 \\
    HSCNet++$^\dagger$ & 76.8 & 67.4 & 61.5 & 78.5 & 67.2 & 73.5 & 56.0 & 68.7  \\
    \midrule
    HSCNet++ & \textbf{77.1} & \textbf{69.6} & \textbf{67.1} & \textbf{79.5} & \textbf{70.1} & \textbf{75.5} & \textbf{56.1} & \textbf{70.7} \\
    \bottomrule
    \end{tabular}
    \caption{
    Sub-region prediction accuracy (\%). Results show that HSCNet++ improves classification accuracy at the sub-region level.
    \label{tab:cls_score}
    }
    \end{subtable}
        
    \caption{\textbf{Ablations for HSCNet++.} We analyze the influence of  different design choices of the proposed approach on i7-Scenes.}
\end{table*}

 \noindent{\textbf{Hierarchy and partition granularity.}
The robustness of HSCNet to the label hierarchy hyperparameter by varying depth and width are reported in \tab{eval:ll}.
The results show that the performance of our approach is  robust \wrt the choice of these hyperparameters, with a significant drop in performance observed only for the smallest 2-level label hierarchy. Increasing the number of classification layers from 2 is not always beneficial and only brings marginal improvement in 7-Scenes, while increasing the computational costs. We observe the best trade-off for the partition of $25 \times 25$ for both 7-Scenes and $175 \times 25$ for i7-Scenes ($175=7\times25$ due to 7 scenes combined).

\subsection{Ablations: HSCNet++} \label{sec:hscnet++abl}

\noindent\textbf{Impact of internal transformer encoder layers.} 
In this ablation, we remove transformers encoders $t_r$ and $t_s$, while only $t_{3D}$ remains. This variant is denoted by HSCNet++$^\dagger$ and Table~\ref{tbl:ablation_number_of_params} shows a small to noticeable drop in all cases.

To factor out the impact of multi-headed attention (MHA) layers, we report results in Table~\ref{tbl:ablation_number_of_params}, which shows that increasing the number of MHA layers in HSCNet++$^\dagger$ does not lead to performance improvement. It is worth mentioning that HSCNet++$^\dagger$ with 8 MHA layers has 2 million more parameters than HSCNet++.
Our intuition is that this happens due to the improvement of predictions at coarse levels of the network. 
To test the above hypothesis, we compute the accuracy of the sub-region predictions.
For each valid pixel in a query image 
, this metric evaluates whether the valid pixel 
is correctly classified. Results in~ Table~\ref{tab:cls_score} show that adding transformers at classification branches helps to improve the label classification accuracy. However, the sub-region prediction accuracy does not always correlate with the localization performance.
This can be attributed to RANSAC-based filtering of final 3D scene coordinates for camera pose estimation. That is, incorrect 3D scene predictions due to erroneous sub-region predictions can be detected as outliers by RANSAC.

\noindent \textbf{Impact of positional encoding.} 
We compare the proposed way of providing region (position) information to the transformer blocks 
with the classical positional encoding used in transformers. 
As label encoding is an inherent part of HSCNet, for a direct comparison with positional encoding, we additionally add the positional encoding right before the transformer block and perform experiments on i7-Scenes. 
Results presented in Table~\ref{tbl:ablation_positional_encoding} show that with the additional position encoding the results noticeably drop.

\subsection{Results for HSCNet++(S)}
\label{sec:hscnet++}

\begin{figure*}[t!]
\begin{center}
\includegraphics[width=0.99\textwidth]{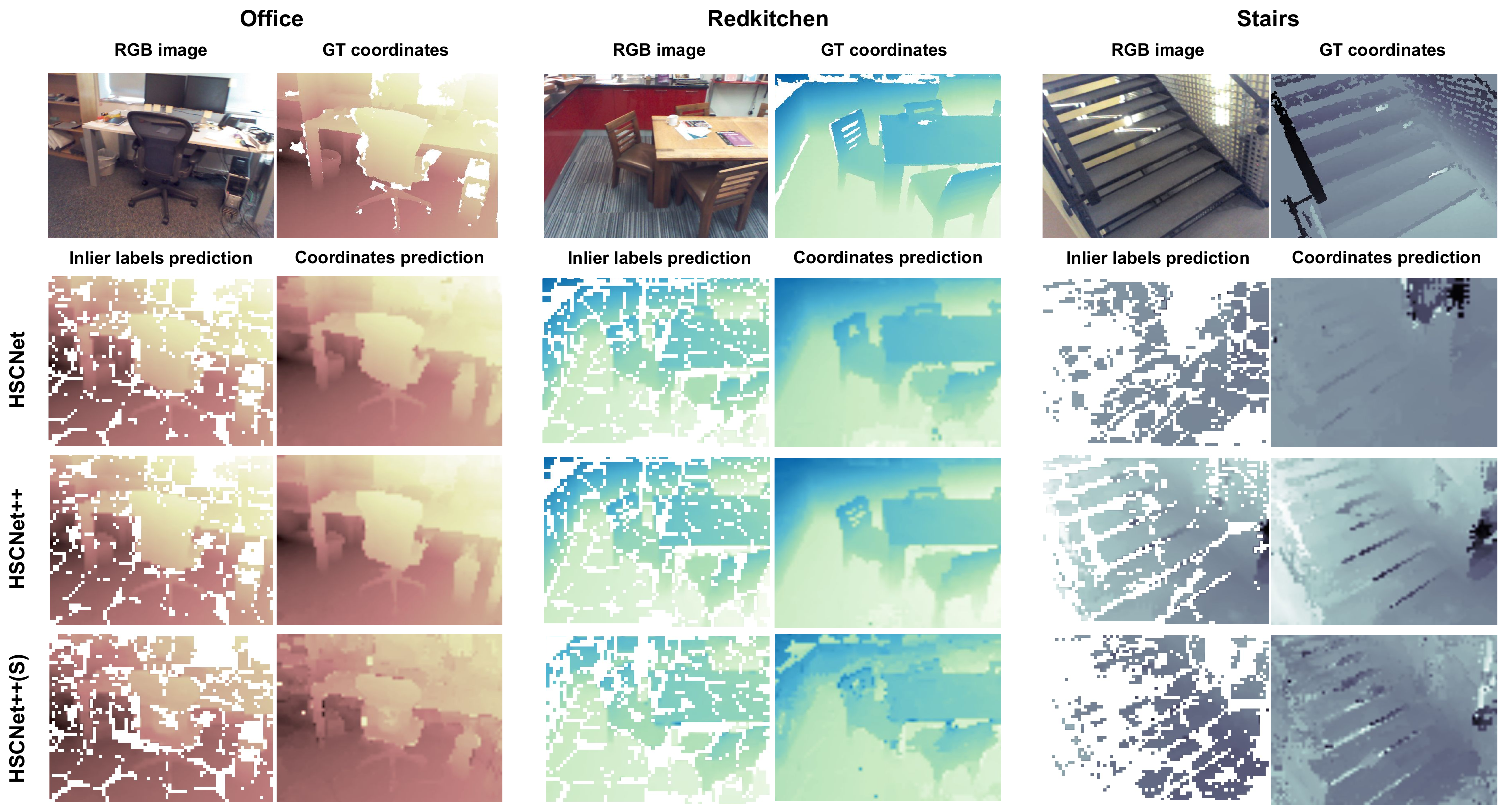}
\end{center}
\caption{
\textbf{Scene coordinates visualization on i7-Scenes.} We visualize the scene coordinate predictions for three test images with HSCNet, HSCNet++, and HSCNet++(S) on i7-Scenes. The XYZ coordinates are mapped to the heatmap, and the ground truth scene coordinates are computed from the depth maps. For each image, the left column is the correct predicted label and the right column is the predicted scene coordinates.}
\label{fig:pred_viz}
\end{figure*}

\begin{figure*}[t]
\begin{center}
\vspace{15pt}
\includegraphics[width=1\textwidth]{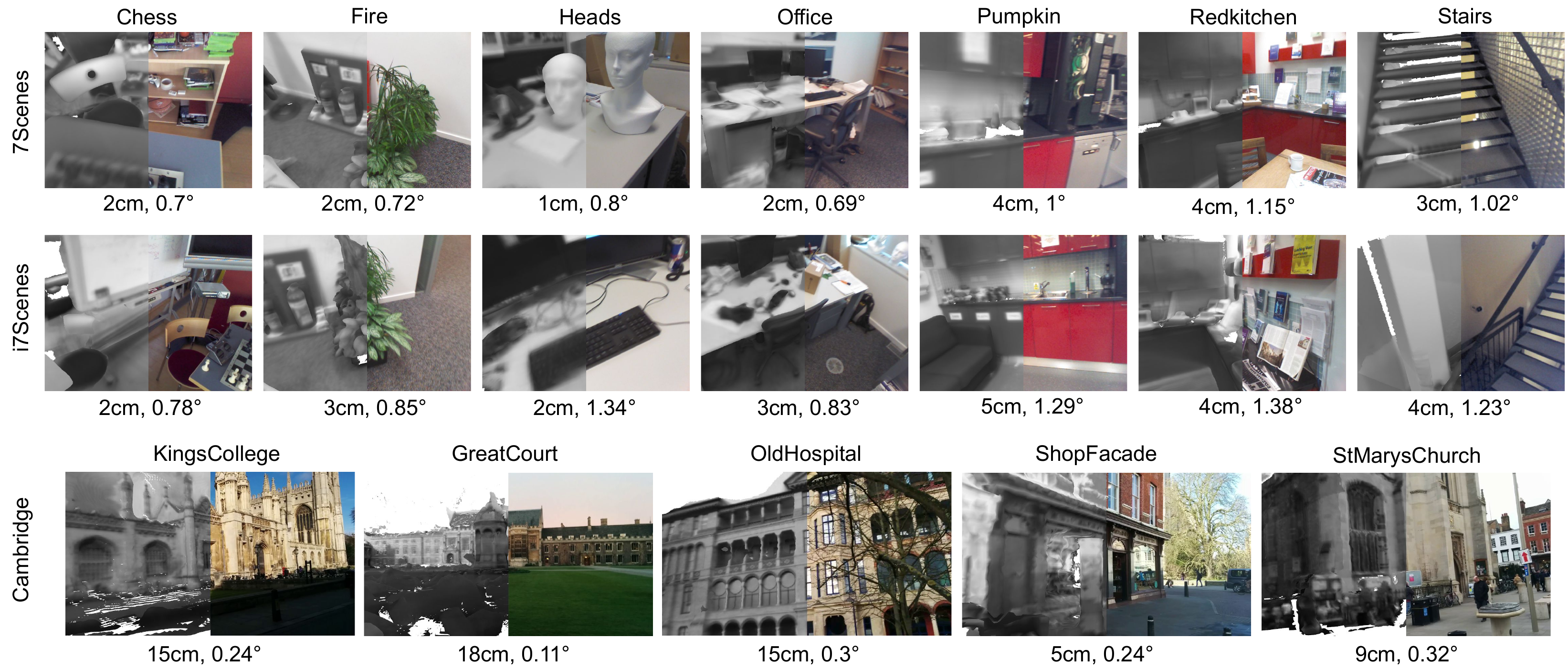}
\end{center}
\caption{
\textbf{Median Error for HSCNet++(S)}. We show the frames with median pose estimation error in each scene and visualize the accuracy by overlaying the query image (right) with a rendered image (left, grayscale) using the estimated pose and the ground truth 3D model. }

\label{fig:pred_viz}
\end{figure*}

We now present results for HSCNet with sparse supervision and study the pseudo-labeling and loss functions in detail. We donate it as HSCNet++(S).
For indoor scenes, we synthetically sparsify dense coordinates using sparse SIFT-based SfM reconstruction. That is, we select the subset of dense 3D coordinates whose 2D reprojections (pixel locations) are also registered in the SfM reconstruction. For the outdoor Cambridge dataset, we directly obtain the keypoints of training images from the provided SfM models. 

The localization performance on 7-Scenes, i7-Scenes, and Cambridge datasets is provided in Fig~\ref{fig:pred_viz} and Table~\ref{tbl:hscnet++s}. 
Results show that even with sparse coordinate supervision, HSCNet++(S) achieves competitive results on 7-Scenes with respect to the dense counterpart, even outperforming HSCNet. On the more challenging combined scene setup of i7-Scenes, HSCNet++(S) lacks by 10\% indicating a further requirement for future research in this direction. However, on the outdoor dataset Cambridge Landmarks, where only sparse coordinate data is available in most cases, 
HSCNet++(S) outperforms HSCNet and HSCNet++, which are trained on MVS-densified~\cite{Brachmann_2018_CVPR, schoenberger2016mvs, li2020hscnet} data, by a large margin. 
It demonstrates the effectiveness of our label propagation and supports our hypothesis that noisy dense ground truth from MVS harms the training process. The largest improvement is observed on Kings College, Great Court and Old Hospital with median pose errors ($cm/\circ$) of $15/0.24$, $18/0.11$ and $15/0.30$ respectively (\cf Table~\ref{tbl:cam_dense}). On average median pose error, HSCNet++ (S) outperforms PixLoc (15/0.25), VSNet (13.6/0.24) and DSAC* (20.6/0.34).

\begin{table}[t!]
    \centering \fontsize{8}{9}\selectfont
	\renewcommand{\tabcolsep}{7pt}
		\begin{tabular}{l | ccc}
        \toprule
        \multirow{3}{*}{Method} & \multicolumn{3}{c}{Localization}\\
        & \multicolumn{2}{c}{Accuracy (\%)~$\uparrow$} & Error (cm/$^\circ$)~$\downarrow$\\
        & 7-Scenes & i7-Scenes & Cambridge  \\
        \midrule
        HSCNet   & 84.8 & 83.3 &  16.0 / 0.28 \\
        HSCNet++ & \bfseries 88.7 & \bfseries 88.3 & 18.6 / 0.28 \\
        HSCNet++(S)    & 85.2 &  78.5 & \textbf{12.4 / 0.24} \\
        \bottomrule
        \end{tabular}
	\caption{\textbf{HSCNet++(S) results}. The table presents average localization accuracy (\%) under $5cm/5^\circ$ and average median pose error (cm/$\circ$) of HSCNet++(S) and dense counterparts on 7-Scenes, i7-Scenes and Cambridge.}
	\label{tbl:hscnet++s}
\end{table}

\begin{table*}[t!]
\vspace{10pt}
	\centering \fontsize{8}{9}\selectfont
	\renewcommand{\tabcolsep}{9pt}
		\begin{tabular}{c l | c c c c c c c | c}
		\toprule
		 & {Method} & Chess & Fire & Heads & Office & Pumpkin & Kitchen & Stairs & {Average} \\
		 \midrule
		  \multirow{5}{*}{\rotatebox[origin=c]{90}{Error}} 
		  & {HSCNet++(S)} & 2 / 0.70 & 2 / 0.72 & 1 / 0.8 & 2 / 0.69 & 4 / 1.00 & 4 / 1.15 & 3 / 1.02 & - \\		  
		  & {\hspace{5pt} w/o LP} & 3 / 0.86 & 3 / 0.91 & 3 / 1.47 & 5 / 1.15 & 6 / 1.37 & 5 / 1.39 & 7 / 1.91 & - \\
		  & {\hspace{5pt} w/o Txf} & 2 / 0.70 & 2 / 0.94 & 1 / 0.76 & 3 / 0.78 & 4 / 1.12 & 4 / 1.2 & 3 / 1.01 & - \\		  
		  & {\hspace{5pt} w/o SCE} & 2 / 0.75 & 2 / 0.77 & 1 / 0.85 & 3 / 0.71 & 4 / 1.04 & 4 / 1.14 & 4 / 1.03 & - \\
		  & {\hspace{5pt} w/o Rep} & 2 / 0.70 & 2 / 0.80 & 1 / 0.93 & 3 / 0.81 & 4 / 1.09 & 4 / 1.35 & 5 / 1.32 & - \\
		  \midrule
		  \multirow{5}{*}{\rotatebox[origin=c]{90}{Accuracy}} 
		  & {HSCNet++(S)} & 98.1 & 97.0 & 98.8 & 88.2 & 65.1 & 72.9 & 76.6 & 85.2 \\		  
		  & {\hspace{5pt} w/o LP} & 86.0 & 81.1 & 85.4 & 56.0 & 39.4 & 49.6 & 36.2 & 62.0 \\
		  & {\hspace{5pt} w/o Txf} & 97.3 & 98.8 & 99.6 & 85.6 & 59.4 & 64.4 & 80.5 & 83.7 \\		  
		  & {\hspace{5pt} w/o SCE} & 97.6 & 96.2 & 96.5 & 84.2 & 64.1 & 70.1 & 73.2 & 83.1 \\
		  & {\hspace{5pt} w/o Rep} & 97.5 & 98.2 & 96.8 & 80.2 & 62.8 & 64.8 & 55.0 & 79.3 \\
		  \bottomrule
		  \end{tabular}
\caption{
\textbf{Ablations for HSCNet++(S)} The results of HSCNet++(S) and various variants are presented, the table shows the median translation and rotation errors (Error) and localization accuracy (Accuracy) under 5cm/5$^\circ$. 
\label{tab:sparse_abla}
}	
\end{table*}

\noindent\textbf{Component Ablations.} We formulate ablations on 7-Scenes to examine the components in the proposed HSCNet++(S). We first train the model without the proposed label propagation i.e only with sparse keypoint pixels only as the baseline. Then, for the HSCNet++(S), we present three variants by removing each component - transformers, symmetric cross-entropy and reprojection loss in HSCNet++(S) as shown in Table~\ref{tab:sparse_abla}. The baseline achieves only 62.0\% on average accuracy which is significantly worse than our result (85.2\%). Variants \textit{w/o} \textbf{Txf}, \textbf{SCE} and \textbf{Rep} models show worse performance compared to HSCNet++(S) on average. Results demonstrate that the synergy of individual components leads the superior results.

\begin{table}[h]
\centering \fontsize{8}{9}\selectfont
	\renewcommand{\tabcolsep}{7pt}
		\begin{tabular}{l | c c c | c c  }
		\toprule
		 \multirowcell{3}{Methods} & \multicolumn{5}{c}{Scenes} \\
		 & \multicolumn{3}{c}{Red Kitchen} & \multicolumn{2}{c}{GreatCourt} \\
 		 & $\bm{t}$, cm & $\bm{r}$, $^\circ$ & Accuracy, \% & $\bm{t}$, cm & $\bm{r}$, $^\circ$ \\
		\midrule
		$z = 0$ & 6 & 1.37 & 65.5 & 32 & 0.28 \\
		$z = 3$ & 4 & 1.14 & 70.3 & 18 & 0.11 \\
		$z = 5$ & 4 & 1.15 & 72.9 & 18 & 0.11 \\
		$z = 7$ & 4 & 1.12 & 71.7 & 21 & 0.14 \\
		$z = 9$ & 3 & 1.12 & 73.0 & 35 & 0.20 \\
        \bottomrule
		\end{tabular}
		\caption{\textbf{Impact of $z$ on pose estimation} We report the pose estimation results (median errors and accuracy) on Red Kitchen and Great Court with different neighborhood size}\label{tab:dsize}
\end{table}

\begin{figure}[t!]
\hspace{-15pt}
    \includegraphics[scale = 0.4]{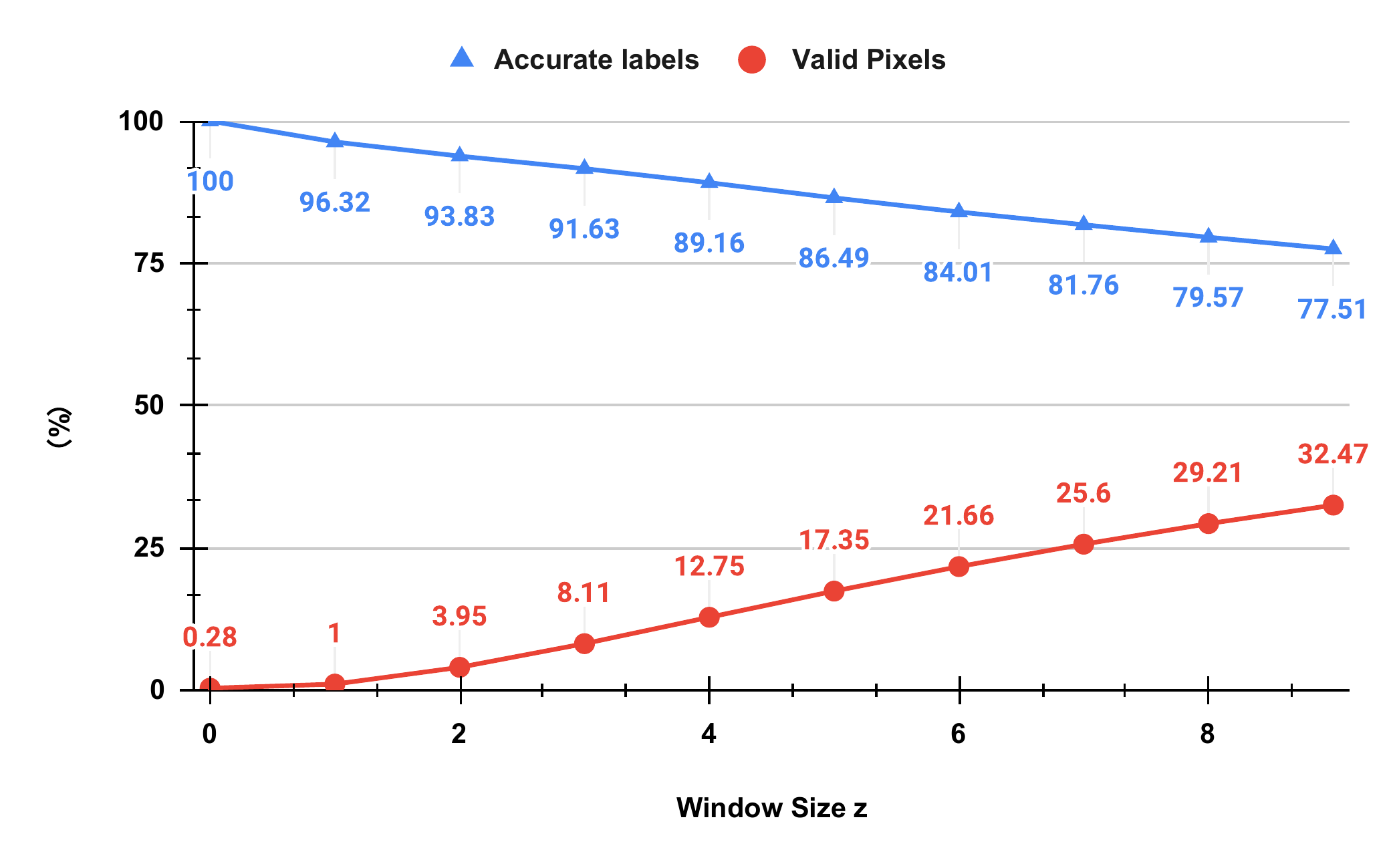}
    \caption{\textbf{Impact of neighborhood size $z$.} The percentage of accurate labels and valid pixels change with the increasing of neighborhood size $z$.}
    \label{fig:z_size}

\end{figure}

\noindent\textbf{Impact of LP Neighborhood Size. } In this section, we analyze the impact of the LP neighborhood size, z. We vary the neighborhood size $z$ range from $0 \rightarrow 9$ on RedKitchen as ablation, and the results are reported in Fig~\ref{fig:z_size} and Table~\ref{tab:dsize}. 
Fig~\ref{fig:z_size} shows that increasing the size of z, also increases pseudo-label noise shown by a decrease in the percentage of accurate labels. For \eg when $z = 5$ the fraction of noisy labels is 15\%. Results in Table~\ref{tab:dsize} 
shows that there is a trade-off between increasing $z$, and camera localization accuracy. This effect is more pronounced in outdoor scene, Great Court from Cambridge dataset, where increasing $z$ from $0 \rightarrow 5$ reduces median pose error (\textbf{t/r}) from $32/0.28 \rightarrow 18/0.11 $. But increasing $z$ further from $10 \rightarrow 18$ increases median pose error from $18/0.11 \rightarrow 35/0.2$. Limiting spatial proximity of pseudo-labels to initial sparse labels seems a suitable choice.

\subsection{Model Capacity and Efficiency}

\begin{table}[t!]
\centering 
\fontsize{6}{7}\selectfont
\renewcommand{\tabcolsep}{2pt}
    \begin{tabular}{l|cc|cc}
    \toprule
    \multirow{2}{*}{Dataset} & \multicolumn{2}{c}{7-Scenes}&\multicolumn{2}{c}{i7-Scenes} \\
    & HSCNet & HSCNet++ & HSCNet   & HSCNet++           \\
    \midrule
Model Size, Mb   & 147.9 & 84.5 & 163 & 113.5               \\
Training time, ms/iter    & $\sim$125 & $\sim$89 & $\sim$135 & $\sim$133 \\
Inference time, ms/query    & \multicolumn{4}{c}{${\sim}$85--130}  \\
\bottomrule
\end{tabular}

\caption{\textbf{Comparison of the model capacity and runtime}. We compare the statistics of the model of HSCNet and HSCNet++, we provide the results on the same software and hardware setting.
\label{tab:running}
}

\end{table}

\label{sec:eff}

\noindent\textbf{Model Capacity.} As mentioned in Sec.~\ref{sec:hscnet++(d)}, we prune some heavy convolution layers compared to HSCNet. To demonstrate the efficiency of this setting, Table~\ref{tab:running} reports the model size of HSCNet and HSCNet++ on 7-Scenes and i7-Scenes.  Our method has a memory footprint reduction of 43\% compared to HSCNet on the individual scene training and 30\% reduction on the combined scenes.

\noindent\textbf{Runtime.} For a fair comparison of the running time, we run all the experiments on NVIDIA GeForce RTX 2080 Ti GPU and AMD Ryzen Threadripper 2950x CPU. It takes $\sim$7.4 h for 300k iterations on individual scene training for HSCNet++ and $\sim$10.4 h on HSCNet with the same setting. We show the approximate training time for one iteration in Table~\ref{tab:running}. It is clear that HSCNet++ has a smaller memory footprint and faster training time while offering higher accuracy. We also notice that the training time grows with the number of multi-head attention layer increases. 

We have not observed a clear difference between the two methods in the inference running time. The running time varies from around 85 ms to 130 ms to localize one image. This is mainly dependent on the accuracy of predicted 2D-3D correspondences fed into the RANSAC-PnP loop.

\section{Conclusion}
We have propsoed a novel hierarchical coarse-to-fine approach for scene coordinate prediction. The network benefits from FiLM-like conditioning of coarse region predictions for better scene coordinate prediction. Experimentally we demonstrate that both hierarchical and prediction conditioning are required for improvement. The method is extended to handle sparse labels using the proposed pseudo-labeling approach. Adaptation of symmetric cross-entropy and reprojection losses provides robustness to pseudo-label noise. We also show that synergy of each component proposed in this work is needed for best performance.

Results show that the proposed hierarchical scene coordinate network is more accurate than previous regression only approaches for single-image RGB localization. The proposed method is also more scalable as shown by results on three indoor datasets.
In addition, the proposed method is extended to handle sparse labels using less costly methods than existing methods and obtaining better results on outdoor scenes.

\section{Acknowledgments}
This work was supported by the Academy of Finland (grant No. 327911, 353138) and Junior Star GACR (grant No. GM 21-28830M). We acknowledge the computational resources provided by the Aalto Science-IT project, CSC-IT Center for Science, Finland and OP VVV funded project CZ.02.1.01/ 0.0/0.0/16\_019/0000765 ``Research Center for Informatics''. We thank Dr. Jakob Verbeek for contributing the HSCNet publication.


{\small
\bibliographystyle{spmpsci}
\bibliography{main}

\begin{thebibliography}{10}
\providecommand{\url}[1]{{#1}}
\providecommand{\urlprefix}{URL }
\expandafter\ifx\csname urlstyle\endcsname\relax
  \providecommand{\doi}[1]{DOI~\discretionary{}{}{}#1}\else
  \providecommand{\doi}{DOI~\discretionary{}{}{}\begingroup
  \urlstyle{rm}\Url}\fi

\bibitem{NetVLAD}
Arandjelovi\'c, R., Gronat, P., Torii, A., Pajdla, T., Sivic, J.: {NetVLAD}:
  {CNN} architecture for weakly supervised place recognition.
\newblock In: \emph{Proceedings of the IEEE/CVF Conference on Computer Vision
  and Pattern Recognition (CVPR)}, pp. 5297--5307 (2016)

\bibitem{Balntas_2018_ECCV}
Balntas, V., Li, S., Prisacariu, V.: {RelocNet}: Continuous metric learning
  relocalisation using neural nets.
\newblock In: \emph{Proceedings of the European Conference on Computer Vision
  (ECCV)}, pp. 751--767. Springer International Publishing (2018)

\bibitem{Balntas2016TFeat}
Balntas, V., Riba, E., Ponsa, D., Mikolajczyk, K.: Learning local feature
  descriptors with triplets and shallow convolutional neural networks.
\newblock In: \emph{Proceedings of the British Machine Vision Conference
  (BMVC)} (2016)

\bibitem{Bay2006SURF}
Bay, H., Tuytelaars, T., Van~Gool, L.: {SURF}: Speeded up robust features.
\newblock In: \emph{Proceedings of the European Conference on Computer Vision
  (ECCV)}, pp. 404--417. Springer International Publishing (2006)

\bibitem{brachmann2021limits}
Brachmann, E., Humenberger, M., Rother, C., Sattler, T.: On the limits of
  pseudo ground truth in visual camera re-localisation.
\newblock In: \emph{Proceedings of the IEEE/CVF International Conference on
  Computer Vision (ICCV)}, pp. 6218--6228 (2021)

\bibitem{Brachmann_2017_CVPR}
Brachmann, E., Krull, A., Nowozin, S., Shotton, J., Michel, F., Gumhold, S.,
  Rother, C.: {DSAC} - {D}ifferentiable {RANSAC} for camera localization.
\newblock In: \emph{Proceedings of the IEEE/CVF Conference on Computer Vision
  and Pattern Recognition (CVPR)}, pp. 6684--6692 (2017)

\bibitem{BrachmannMKYGR16}
Brachmann, E., Michel, F., Krull, A., Yang, M.Y., Gumhold, S., Rother, C.:
  Uncertainty-driven {6D} pose estimation of objects and scenes from a single
  {RGB} image.
\newblock In: \emph{Proceedings of the IEEE/CVF Conference on Computer Vision
  and Pattern Recognition (CVPR)}, pp. 3364--3372 (2016)

\bibitem{Brachmann_2018_CVPR}
Brachmann, E., Rother, C.: Learning less is more - {6D} camera localization via
  {3D} surface regression.
\newblock In: \emph{Proceedings of the IEEE/CVF Conference on Computer Vision
  and Pattern Recognition (CVPR)}, pp. 4654--4662 (2018)

\bibitem{Brachmann2019SampleConsensus}
Brachmann, E., Rother, C.: Expert sample consensus applied to camera
  re-localization.
\newblock In: \emph{Proceedings of the IEEE/CVF International Conference on
  Computer Vision (ICCV)}, pp. 7524--7533 (2019)

\bibitem{Brachmann_2019_ICCV_NG}
Brachmann, E., Rother, C.: Neural-guided {RANSAC}: Learning where to sample
  model hypotheses.
\newblock In: \emph{Proceedings of the IEEE/CVF International Conference on
  Computer Vision (ICCV)}, pp. 4322--4331 (2019)

\bibitem{brachmann2020visual}
Brachmann, E., Rother, C.: Visual camera re-localization from {RGB} and {RGB-D}
  images using {DSAC}.
\newblock \emph{IEEE Transactions on Pattern Analysis and Machine Intelligence}
  \textbf{44}(9), 5847--5865 (2021)

\bibitem{mapnet2018}
Brahmbhatt, S., Gu, J., Kim, K., Hays, J., Kautz, J.: Geometry-aware learning
  of maps for camera localization.
\newblock In: \emph{Proceedings of the IEEE/CVF Conference on Computer Vision
  and Pattern Recognition (CVPR)}, pp. 2616--2625 (2018)

\bibitem{Budvytis2019}
Budvytis, I., Teichmann, M., Vojir, T., Cipolla, R.: Large scale joint semantic
  re-localisation and scene understanding via globally unique instance
  coordinate regression.
\newblock In: \emph{Proceedings of the British Machine Vision Conference
  (BMVC)} (2019)

\bibitem{bui2018scene}
Bui, M., Albarqouni, S., Ilic, S., Navab, N.: Scene coordinate and
  correspondence learning for image-based localization.
\newblock In: \emph{Proceedings of the British Machine Vision Conference
  (BMVC)} (2018)

\bibitem{Calonder2019Brief}
Calonder, M., Lepetit, V., Strecha, C., Fua, P.: {BRIEF}: Binary robust
  independent elementary features.
\newblock In: \emph{Proceedings of the European Conference on Computer Vision
  (ECCV)}, pp. 778--792. Springer Berlin Heidelberg (2010)

\bibitem{Cavallari_corr_19}
Cavallari, T., Bertinetto, L., Mukhoti, J., Torr, P., Golodetz, S.: Let's take
  this online: Adapting scene coordinate regression network predictions for
  online {RGB-D} camera relocalisation.
\newblock In: \emph{International Conference on 3D Vision (3DV)}, pp. 564--573
  (2019)

\bibitem{cavallari2019real}
Cavallari, T., Golodetz, S., Lord, N., Valentin, J., Prisacariu, V.,
  Di~Stefano, L., Torr, P.H.: Real-time {RGB-D} camera pose estimation in novel
  scenes using a relocalisation cascade.
\newblock \emph{IEEE Transactions on Pattern Analysis and Machine Intelligence}
  \textbf{42}(10), 2465--2477 (2020)

\bibitem{CavallariGLVST17}
Cavallari, T., Golodetz, S., Lord, N.A., Valentin, J., Di~Stefano, L., Torr,
  P.H.: On-the-fly adaptation of regression forests for online camera
  relocalisation.
\newblock In: \emph{Proceedings of the IEEE/CVF Conference on Computer Vision
  and Pattern Recognition (CVPR)}, pp. 4457--4466 (2017)

\bibitem{Chen2022DFnet}
Chen, S., Li, X., Wang, Z., Prisacariu, V.: Dfnet: Enhance absolute pose
  regression with direct feature matching.
\newblock In: \emph{Proceedings of the European Conference on Computer Vision
  (ECCV)}, pp. 1--17. Springer Nature Switzerland (2022)

\bibitem{Chen2021DirectPoseNet}
Chen, S., Wang, Z., Prisacariu, V.: Direct-posenet: Absolute pose regression
  with photometric consistency.
\newblock In: \emph{International Conference on 3D Vision (3DV)}, pp.
  1175--1185 (2021)

\bibitem{DeTone_2018_CVPR_Workshops}
DeTone, D., Malisiewicz, T., Rabinovich, A.: {SuperPoint}: Self-supervised
  interest point detection and description.
\newblock In: \emph{Proceedings of the IEEE Conference on Computer Vision and
  Pattern Recognition (CVPR) Workshops}, pp. 337--349 (2018)

\bibitem{Ding_2019_ICCV}
Ding, M., Wang, Z., Sun, J., Shi, J., Luo, P.: {CamNet}: Coarse-to-fine
  retrieval for camera re-localization.
\newblock In: \emph{Proceedings of the IEEE/CVF International Conference on
  Computer Vision (ICCV)}, pp. 2871--2880 (2019)

\bibitem{Dusmanu2019CVPR}
Dusmanu, M., Rocco, I., Pajdla, T., Pollefeys, M., Sivic, J., Torii, A.,
  Sattler, T.: {D2-Net}: A trainable {CNN} for joint detection and description
  of local features.
\newblock In: \emph{Proceedings of the IEEE/CVF Conference on Computer Vision
  and Pattern Recognition (CVPR)}, pp. 8092--8101 (2019)

\bibitem{RANSAC}
Fischler, M.A., Bolles, R.C.: Random sample consensus: A paradigm for model
  fitting with applications to image analysis and automated cartography.
\newblock \emph{Communications of the ACM} \textbf{24}(6), 381--395 (1981)

\bibitem{9437699}
Guan, P., Cao, Z., Yu, J., Zhou, C., Tan, M.: Scene coordinate regression
  network with global context-guided spatial feature transformation for visual
  relocalization.
\newblock \emph{IEEE Robotics and Automation Letters} \textbf{6}(3), 5737--5744
  (2021)

\bibitem{Guzman-RiveraKGSSFI14}
Guzm{\'{a}}n{-}Rivera, A., Kohli, P., Glocker, B., Shotton, J., Sharp, T.,
  Fitzgibbon, A.W., Izadi, S.: Multi-output learning for camera relocalization.
\newblock In: \emph{Proceedings of the IEEE/CVF Conference on Computer Vision
  and Pattern Recognition (CVPR)}, pp. 1114--1121 (2014)

\bibitem{Han2015MatchNet}
Han, X., Leung, T., Jia, Y., Sukthankar, R., Berg, A.C.: Matchnet: Unifying
  feature and metric learning for patch-based matching.
\newblock In: \emph{Proceedings of the IEEE/CVF Conference on Computer Vision
  and Pattern Recognition (CVPR)}, pp. 3279--3286 (2015)

\bibitem{huang2021vs}
Huang, Z., Zhou, H., Li, Y., Yang, B., Xu, Y., Zhou, X., Bao, H., Zhang, G.,
  Li, H.: {VS-Net}: Voting with segmentation for visual localization.
\newblock In: \emph{Proceedings of the IEEE/CVF Conference on Computer Vision
  and Pattern Recognition (CVPR)}, pp. 6101--6111 (2021)

\bibitem{Jiang2021COTR}
Jiang, W., Trulls, E., Hosang, J., Tagliasacchi, A., Yi, K.M.: {COTR}:
  Correspondence transformer for matching across images.
\newblock In: \emph{Proceedings of the IEEE/CVF International Conference on
  Computer Vision (ICCV)}, pp. 6207--6217 (2021)

\bibitem{katharopoulos2020transformers}
Katharopoulos, A., Vyas, A., Pappas, N., Fleuret, F.: Transformers are rnns:
  Fast autoregressive transformers with linear attention.
\newblock In: \emph{Proceedings of the 37th International Conference on Machine
  Learning (ICML)}, pp. 5156--5165. JMLR (2020)

\bibitem{KendallC15bay}
Kendall, A., Cipolla, R.: Modelling uncertainty in deep learning for camera
  relocalization.
\newblock In: \emph{Proceedings of the IEEE International Conference on
  Robotics and Automation (ICRA)}, pp. 4762--4769 (2016)

\bibitem{Kendall_2017_CVPR}
Kendall, A., Cipolla, R.: Geometric loss functions for camera pose regression
  with deep learning.
\newblock In: \emph{Proceedings of the IEEE/CVF Conference on Computer Vision
  and Pattern Recognition (CVPR)}, pp. 5974--5983 (2017)

\bibitem{Kendall2018Uncertainty}
Kendall, A., Gal, Y., Cipolla, R.: Multi-task learning using uncertainty to
  weigh losses for scene geometry and semantics.
\newblock In: \emph{Proceedings of the IEEE/CVF Conference on Computer Vision
  and Pattern Recognition (CVPR)}, pp. 7482--7491 (2018)

\bibitem{kendall2015convolutional}
Kendall, A., Grimes, M., Cipolla, R.: {PoseNet}: A convolutional network for
  real-time {6-DoF} camera relocalization.
\newblock In: \emph{Proceedings of the IEEE/CVF International Conference on
  Computer Vision (ICCV)}, pp. 2938--2946 (2015)

\bibitem{LaskarMKK17}
Laskar, Z., Melekhov, I., Kalia, S., Kannala, J.: Camera relocalization by
  computing pairwise relative poses using convolutional neural network.
\newblock In: \emph{Proceedings of the IEEE/CVF International Conference on
  Computer Vision (ICCV) Workshops}, pp. 929--938 (2017)

\bibitem{li2020hscnet}
Li, X., Wang, S., Zhao, Y., Verbeek, J., Kannala, J.: Hierarchical scene
  coordinate classification and regression for visual localization.
\newblock In: \emph{Proceedings of the IEEE/CVF Conference on Computer Vision
  and Pattern Recognition (CVPR)}, pp. 11,983--11,992 (2020)

\bibitem{Li2018}
Li, X., Ylioinas, J., Kannala, J.: Full-frame scene coordinate regression for
  image-based localization.
\newblock In: \emph{Proceedings of Robotics: Science and Systems (RSS)} (2018)

\bibitem{Li_Ylioinas_Verbeek_Kannala_2018}
Li, X., Ylioinas, J., Verbeek, J., Kannala, J.: Scene coordinate regression
  with angle-based reprojection loss for camera relocalization.
\newblock In: \emph{Proceedings of the European Conference on Computer Vision
  (ECCV) Workshops}, pp. 229--245. Springer International Publishing (2018)

\bibitem{SIFT}
Lowe, D.G.: Distinctive image features from scale-invariant keypoints.
\newblock \emph{International Journal of Computer Vision} \textbf{60}(2),
  91--110 (2004)

\bibitem{luo2019contextdesc}
Luo, Z., Shen, T., Zhou, L., Zhang, J., Yao, Y., Li, S., Fang, T., Quan, L.:
  Contextdesc: Local descriptor augmentation with cross-modality context.
\newblock In: \emph{Proceedings of the IEEE/CVF Conference on Computer Vision
  and Pattern Recognition (CVPR)}, pp. 2527--2536 (2019)

\bibitem{Massiceti2017}
Massiceti, D., Krull, A., Brachmann, E., Rother, C., Torr, P.H.: Random forests
  versus neural networks - {What's} best for camera localization{?}
\newblock In: \emph{Proceedings of the IEEE International Conference on
  Robotics and Automation (ICRA)}, pp. 5118--5125 (2017)

\bibitem{Melekhov2020Stylization}
Melekhov, I., Brostow, G.J., Kannala, J., Turmukhambetov, D.: Image stylization
  for robust features.
\newblock ArXiv preprint arXiv:2008.06959  (2020)

\bibitem{Melekhov2017PatchMatch}
Melekhov, I., Kannala, J., Rahtu, E.: Image patch matching using convolutional
  descriptors with euclidean distance.
\newblock In: \emph{Proceedings of the Asian Conference on Computer Vision
  (ACCV) Workshops}, pp. 638--653. springer (2017)

\bibitem{Melekhov2021hndesc}
Melekhov, I., Laskar, Z., Li, X., Wang, S., Juho, K.: Digging into
  self-supervised learning of feature descriptors.
\newblock In: \emph{International Conference on 3D Vision (3DV)}, pp.
  1144--1155 (2021)

\bibitem{MelekhovYKR17}
Melekhov, I., Ylioinas, J., Kannala, J., Rahtu, E.: Image-based localization
  using hourglass networks.
\newblock In: \emph{Proceedings of the IEEE/CVF International Conference on
  Computer Vision (ICCV) Workshops}, pp. 879--886 (2017)

\bibitem{meng2017backtracking}
Meng, L., Chen, J., Tung, F., Little, J.J., Valentin, J., de~Silva, C.W.:
  Backtracking regression forests for accurate camera relocalization.
\newblock In: \emph{Proceedings of the IEEE/RSJ International Conference on
  Intelligent Robots and Systems (IROS)}, pp. 6886--6893 (2017)

\bibitem{meng2018exploiting}
Meng, L., Tung, F., Little, J.J., Valentin, J., de~Silva, C.W.: Exploiting
  points and lines in regression forests for {RGB-D} camera relocalization.
\newblock In: \emph{Proceedings of the IEEE/RSJ International Conference on
  Intelligent Robots and Systems (IROS)}, pp. 6827--6834 (2018)

\bibitem{Mishchuk2017LocalDescNeigh}
Mishchuk, A., Mishkin, D., Radenovic, F., Matas, J.: Working hard to know your
  neighbor\textquotesingle s margins: Local descriptor learning loss.
\newblock In: \emph{Advances in Neural Information Processing Systems (NIPS)},
  vol.~30, pp. 4826--4837. Curran Associates, Inc. (2017)

\bibitem{moreau2021lens}
Moreau, A., Piasco, N., Tsishkou, D., Stanciulescu, B., de~La~Fortelle, A.:
  {LENS}: Localization enhanced by ne{RF} synthesis.
\newblock In: Annual Conference on Robot Learning (2021)

\bibitem{film}
Perez, E., Strub, F., De~Vries, H., Dumoulin, V., Courville, A.: Film: Visual
  reasoning with a general conditioning layer.
\newblock \emph{Proceedings of the AAAI Conference on Artificial Intelligence}
  \textbf{32}(1), 3942--3951 (2018)

\bibitem{Radenovic2016GEM}
Radenovi{\'c}, F., Tolias, G., Chum, O.: {CNN} image retrieval learns from
  {BoW}: Unsupervised fine-tuning with hard examples.
\newblock In: \emph{Proceedings of the European Conference on Computer Vision
  (ECCV)}, pp. 3--20. Springer International Publishing (2016)

\bibitem{radwan2018vlocnet++}
Radwan, N., Valada, A., Burgard, W.: {VLocNet++}: Deep multitask learning for
  semantic visual localization and odometry.
\newblock \emph{IEEE Robotics and Automation Letters} \textbf{3}(4), 4407--4414
  (2018)

\bibitem{Revaud2019R2D2}
Revaud, J., De~Souza, C., Humenberger, M., Weinzaepfel, P.: {R2D2}: Reliable
  and repeatable detector and descriptor.
\newblock In: \emph{Advances in Neural Information Processing Systems
  (NeurIPS)}, vol.~32, pp. 12,405--12,415. Curran Associates, Inc. (2019)

\bibitem{rogez17cvpr}
Rogez, G., Weinzaepfel, P., Schmid, C.: {LCR-Net}:
  Localization-classification-regression for human pose.
\newblock In: \emph{Proceedings of the IEEE/CVF Conference on Computer Vision
  and Pattern Recognition (CVPR)}, pp. 3433--3441 (2017)

\bibitem{rogez19pami}
Rogez, G., Weinzaepfel, P., Schmid, C.: {LCR-Net++}: Multi-person {2D} and {3D}
  pose detection in natural images.
\newblock \emph{IEEE Transactions on Pattern Analysis and Machine Intelligence}
  \textbf{42}(5), 1146--1161 (2019)

\bibitem{Rublee2011ORB}
Rublee, E., Rabaud, V., Konolige, K., Bradski, G.R.: {ORB: An efficient
  alternative to SIFT or SURF}.
\newblock In: \emph{Proceedings of the IEEE/CVF International Conference on
  Computer Vision (ICCV)}, pp. 2564--2571 (2011)

\bibitem{saha2018improved}
Saha, S., Varma, G., Jawahar, C.: Improved visual relocalization by discovering
  anchor points.
\newblock In: \emph{Proceedings of the British Machine Vision Conference
  (BMVC)} (2018)

\bibitem{sarlin2019coarse}
Sarlin, P.E., Cadena, C., Siegwart, R., Dymczyk, M.: From coarse to fine:
  Robust hierarchical localization at large scale.
\newblock In: \emph{Proceedings of the IEEE/CVF Conference on Computer Vision
  and Pattern Recognition (CVPR)}, pp. 12,716--12,725 (2019)

\bibitem{sarlin2020superglue}
Sarlin, P.E., DeTone, D., Malisiewicz, T., Rabinovich, A.: Superglue: Learning
  feature matching with graph neural networks.
\newblock In: \emph{Proceedings of the IEEE/CVF Conference on Computer Vision
  and Pattern Recognition (CVPR)}, pp. 4938--4947 (2020)

\bibitem{sarlin2021back}
Sarlin, P.E., Unagar, A., Larsson, M., Germain, H., Toft, C., Larsson, V.,
  Pollefeys, M., Lepetit, V., Hammarstrand, L., Kahl, F., et~al.: Back to the
  feature: Learning robust camera localization from pixels to pose.
\newblock In: \emph{Proceedings of the IEEE/CVF Conference on Computer Vision
  and Pattern Recognition (CVPR)}, pp. 3247--3257 (2021)

\bibitem{sattler2011fast}
Sattler, T., Leibe, B., Kobbelt, L.: Fast image-based localization using direct
  2d-to-3d matching.
\newblock In: \emph{Proceedings of the IEEE/CVF International Conference on
  Computer Vision (ICCV)}, pp. 667--674 (2011)

\bibitem{sattler2012improving}
Sattler, T., Leibe, B., Kobbelt, L.: Improving image-based localization by
  active correspondence search.
\newblock In: \emph{Proceedings of the European Conference on Computer Vision
  (ECCV)}, pp. 752--765. Springer International Publishing (2012)

\bibitem{sattler2016efficient}
Sattler, T., Leibe, B., Kobbelt, L.: Efficient \& effective prioritized
  matching for large-scale image-based localization.
\newblock \emph{IEEE Transactions on Pattern Analysis and Machine Intelligence}
  \textbf{39}(9), 1744--1756 (2016)

\bibitem{Sattler2019}
Sattler, T., Zhou, Q., Pollefeys, M., Leal-Taixe, L.: Understanding the
  limitations of {CNN-based} absolute camera pose regression.
\newblock In: \emph{Proceedings of the IEEE/CVF Conference on Computer Vision
  and Pattern Recognition (CVPR)}, pp. 3302--3312 (2019)

\bibitem{schoenberger2016mvs}
Sch\"{o}nberger, J.L., Zheng, E., Pollefeys, M., Frahm, J.M.: Pixelwise view
  selection for unstructured multi-view stereo.
\newblock In: \emph{Proceedings of the European Conference on Computer Vision
  (ECCV)} (2016)

\bibitem{Shavit2021TransformersLocalization}
Shavit, Y., Ferens, R., Keller, Y.: Learning multi-scene absolute pose
  regression with transformers.
\newblock In: \emph{Proceedings of the IEEE/CVF International Conference on
  Computer Vision (ICCV)}, pp. 2733--2742 (2021)

\bibitem{shavit2022camera}
Shavit, Y., Keller, Y.: Camera pose auto-encoders for improving pose
  regression.
\newblock In: \emph{Proceedings of the European Conference on Computer Vision
  (ECCV)}, pp. 140--157. Springer International Publishing (2022)

\bibitem{SCoRF}
Shotton, J., Glocker, B., Zach, C., Izadi, S., Criminisi, A., Fitzgibbon, A.:
  Scene coordinate regression forests for camera relocalization in {RGB-D}
  images.
\newblock In: \emph{Proceedings of the IEEE/CVF Conference on Computer Vision
  and Pattern Recognition (CVPR)}, pp. 2930--2937 (2013)

\bibitem{Simo-Serra2015DeepDesc}
Simo-Serra, E., Trulls, E., Ferraz, L., Kokkinos, I., Fua, P., Moreno-Noguer,
  F.: Discriminative learning of deep convolutional feature point descriptors.
\newblock In: \emph{Proceedings of the IEEE/CVF International Conference on
  Computer Vision (ICCV)}, pp. 118--126 (2015)

\bibitem{Sun2021LoFTR}
Sun, J., Shen, Z., Wang, Y., Bao, H., Xiaowei, Z.: {LoFTR}: Detector-free local
  feature matching with transformers.
\newblock In: \emph{Proceedings of the IEEE/CVF Conference on Computer Vision
  and Pattern Recognition (CVPR)}, pp. 8922--8931 (2021)

\bibitem{taira2018inloc}
Taira, H., Okutomi, M., Sattler, T., Cimpoi, M., Pollefeys, M., Sivic, J.,
  Pajdla, T., Torii, A.: Inloc: Indoor visual localization with dense matching
  and view synthesis.
\newblock In: \emph{Proceedings of the IEEE/CVF Conference on Computer Vision
  and Pattern Recognition (CVPR)}, pp. 7199--7209 (2018)

\bibitem{l2net2017Tian}
Tian, Y., Fan, B., Wu, F.: L2-net: Deep learning of discriminative patch
  descriptor in euclidean space.
\newblock In: \emph{Proceedings of the IEEE/CVF Conference on Computer Vision
  and Pattern Recognition (CVPR)}, pp. 661--669 (2017)

\bibitem{Tyszkiewicz2020DISK}
Tyszkiewicz, M., Fua, P., Trulls, E.: {DISK}: Learning local features with
  policy.
\newblock In: \emph{Advances in Neural Information Processing Systems
  (NeurIPS)}, vol.~33, pp. 14,254--14,265. Curran Associates, Inc. (2020)

\bibitem{valada2018deep}
Valada, A., Radwan, N., Burgard, W.: Deep auxiliary learning for visual
  localization and odometry.
\newblock In: \emph{Proceedings of the IEEE International Conference on
  Robotics and Automation (ICRA)}, pp. 6939--6946 (2018)

\bibitem{valentin2016learning}
Valentin, J., Dai, A., Nie{\ss}ner, M., Kohli, P., Torr, P., Izadi, S., Keskin,
  C.: Learning to navigate the energy landscape.
\newblock In: \emph{International Conference on 3D Vision (3DV)}, pp. 323--332
  (2016)

\bibitem{ValentinNSFIT15}
Valentin, J., Nie{\ss}ner, M., Shotton, J., Fitzgibbon, A., Izadi, S., Torr,
  P.H.: Exploiting uncertainty in regression forests for accurate camera
  relocalization.
\newblock In: \emph{Proceedings of the IEEE/CVF Conference on Computer Vision
  and Pattern Recognition (CVPR)}, pp. 4400--4408 (2015)

\bibitem{vaswani2017attention}
Vaswani, A., Shazeer, N., Parmar, N., Uszkoreit, J., Jones, L., Gomez, A.N.,
  Kaiser, L.u., Polosukhin, I.: Attention is all you need.
\newblock In: \emph{Advances in Neural Information Processing Systems
  (NeurIPS)}, vol.~30, pp. 5998--6008. Curran Associates, Inc. (2017)

\bibitem{Walch_2017_ICCV}
Walch, F., Hazirbas, C., Leal-Taixe, L., Sattler, T., Hilsenbeck, S., Cremers,
  D.: Image-based localization using {LSTM}s for structured feature
  correlation.
\newblock In: \emph{Proceedings of the IEEE/CVF International Conference on
  Computer Vision (ICCV)}, pp. 627--637 (2017)

\bibitem{Wang2020CAPS}
Wang, Q., Zhou, X., Hariharan, B., Snavely, N.: Learning feature descriptors
  using camera pose supervision.
\newblock In: \emph{Proceedings of the European Conference on Computer Vision
  (ECCV)}, pp. 757--774. Springer International Publishing (2020)

\bibitem{wang2021continual}
Wang, S., Laskar, Z., Melekhov, I., Li, X., Kannala, J.: Continual learning for
  image-based camera localization.
\newblock In: \emph{Proceedings of the IEEE/CVF International Conference on
  Computer Vision (ICCV)}, pp. 3252--3262 (2021)

\bibitem{wang2019symmetric}
Wang, Y., Ma, X., Chen, Z., Luo, Y., Yi, J., Bailey, J.: Symmetric cross
  entropy for robust learning with noisy labels.
\newblock In: \emph{Proceedings of the IEEE/CVF International Conference on
  Computer Vision (ICCV)}, pp. 322--330 (2019)

\bibitem{Weinzaepfel_2019_CVPR}
Weinzaepfel, P., Csurka, G., Cabon, Y., Humenberger, M.: Visual localization by
  learning objects-of-interest dense match regression.
\newblock In: \emph{Proceedings of the IEEE/CVF Conference on Computer Vision
  and Pattern Recognition (CVPR)}, pp. 5634--5643 (2019)

\bibitem{Xue2019LocalSupportsGlobal}
Xue, F., Wang, X., Yan, Z., Wang, Q., Wang, J., Zha, H.: Local supports global:
  Deep camera relocalization with sequence enhancement.
\newblock In: \emph{Proceedings of the IEEE/CVF International Conference on
  Computer Vision (ICCV)}, pp. 2841--2850 (2019)

\bibitem{Xue2020GnnLocalization}
Xue, F., Wu, X., Cai, S., Wang, J.: Learning multi-view camera relocalization
  with graph neural networks.
\newblock In: \emph{Proceedings of the IEEE/CVF Conference on Computer Vision
  and Pattern Recognition (CVPR)}, pp. 11,375--11,384 (2020)

\bibitem{Zagoruyko2015DeepCompare}
Zagoruyko, S., Komodakis, N.: Learning to compare image patches via
  convolutional neural networks.
\newblock In: \emph{Proceedings of the IEEE/CVF Conference on Computer Vision
  and Pattern Recognition (CVPR)}, pp. 4353--4361 (2015)

\bibitem{Zhou2021Patch2Pix}
Zhou, Q., Sattler, T., Leal-Taix{\'e}, L.: {Patch2Pix}: Epipolar-guided
  pixel-level correspondences.
\newblock In: \emph{Proceedings of the IEEE/CVF Conference on Computer Vision
  and Pattern Recognition (CVPR)}, pp. 4669--4678 (2021)

\end{thebibliography}
}

\end{document}